\documentclass{article}

\usepackage[main,final]{neurips_2026}

\usepackage[utf8]{inputenc} 
\usepackage[T1]{fontenc}    
\usepackage{hyperref}       
\usepackage{url}            
\usepackage{booktabs}       
\usepackage{amsfonts}       
\usepackage{nicefrac}       
\usepackage{microtype}      
\usepackage{xcolor}         
\usepackage{amsmath} 
\usepackage{wrapfig}
\usepackage{graphicx} 
\usepackage{xspace}
\usepackage{pifont}
\usepackage{enumitem}
\usepackage{subcaption}
\usepackage[table]{xcolor}
\usepackage{algorithm} 
\usepackage{algorithmic}

\newcommand{\ourmethod}{GenGLAD\xspace} 
\definecolor{firstcolor}{HTML}{148055}
\definecolor{secondcolor}{HTML}{1468a8}
\definecolor{thirdcolor}{HTML}{ec5931}
\definecolor{Gray}{HTML}{e6e6e6}

\newcommand{\first}[1]{\textcolor{firstcolor}{\textbf{\underline{#1}}}}
\newcommand{\second}[1]{\textcolor{secondcolor}{\textbf{\underline{#1}}}}
\newcommand{\third}[1]{\textcolor{thirdcolor}{\textbf{\underline{#1}}}}

\title{Towards Generalist Graph-Level Anomaly Detection}

\author{
  Junjun Pan$^1$, \ Yixin Liu$^1$\thanks{Corresponding Author.}, \ Yu Zheng$^1$, \
  Fuyi Li$^2$, \ Alan Wee-Chung Liew$^1$, \
  Shirui Pan$^1$ \\
  $^1$School of Information and Communication Technology, Griffith University, Australia \\
  $^2$South Australian immunoGENomics Cancer Institute (SAiGENCI),  Adelaide University, Australia \\
  \texttt{junjun.pan@griffithuni.edu.au, fuyi.li@adelaide.edu.au} \\
  \texttt{\{yixin.liu, yu.zheng, a.liew, s.pan\}@griffith.edu.au}
}

\begin{document}
\maketitle
\begin{abstract}
Graph-level anomaly detection (GLAD) aims to identify graphs that deviate significantly from the majority and plays a crucial role in applications such as molecular analysis and fraud detection. While recent advances have achieved competitive performance, existing approaches mainly follow a dataset-specific paradigm, requiring substantial in-domain data and training cost when applied to new application scenarios. This limits their scalability and practicality in real-world applications where labeled data is scarce and diverse domains are continuously emerging.
In this paper, we investigate the problem of generalist GLAD, which seeks to learn a single unified model capable of detecting anomalous graphs across multiple domains with minimal supervision on the target data. This problem introduces key challenges, including learning transferable patterns from heterogeneous graph datasets and effective adaptation under limited supervision. 
To address these challenges, we propose GenGLAD, a generalist GLAD approach that leverages discriminative subgraph extraction as the core mechanism to learn transferable anomaly knowledge and enable efficient adaptation to new domains. 
Specifically, GenGLAD employs a graph neural network-based discriminative subgraph extraction model to learn the anomaly-discriminative substructures that distinguish anomalous and normal graphs across diverse domains. 
During the test-time adaptation phase, multi-dimensional discrepancies between the extracted subgraph and multi-level contextual information are calculated to characterize anomalies from multiple perspectives. For efficient adaptation, a lightweight scoring module is introduced to automatically decide the most informative signals, improving robustness against misleading patterns.
Extensive experiments on diverse benchmark datasets demonstrate that the proposed approach achieves strong performance, superior generalization ability, and improved data efficiency compared to existing methods. 
\end{abstract}


\section{Introduction}
Graphs are a ubiquitous data structure for modeling discrete objects and their relationships, which makes them widely used in fields such as chemistry, biology, and social network analysis~\cite{wu2020comprehensive, zheng2023towards}. Building on this foundation, graph-level anomaly detection (GLAD) aims to identify graph instances that are significantly different from the majority of graphs~\cite{liu2023towards}. Due to its strong capability for solving various real-world challenges, such as identifying toxic chemicals~\cite{aggarwal2010graph}, spotting fraudster networks~\cite{ma2023towards} and safeguarding multi-agent systems~\cite{pan2026explainable}, GLAD has attracted sustained research attention. Recently, graph neural network (GNN)-based methods have become the dominant approach for the GLAD problem, showcasing strong performance compared to traditional kernel-based methods~\cite{liu2023towards,ma2023towards,wang2025unifying}. 

Despite the competitive detection performance they achieved, existing GLAD approaches follow the traditional ``one model per dataset'' paradigm, requiring large amounts of in-domain data to train a dedicated detection model for each dataset. This learning paradigm brings significant limitations for GLAD applications. 
On the one hand, deploying a dataset-specific model requires sufficient in-domain data for model training, which limits its practical applicability in domains where data are scarce or difficult to collect. On the other hand, training a dataset-specific model is usually computationally expensive, involving high time and hardware costs, making it inefficient for large-scale databases.

\begin{figure*}
\vspace{-2mm}
  \centering
  \begin{subfigure}{0.73\linewidth}
    \centering
    \includegraphics[height=2.6cm]{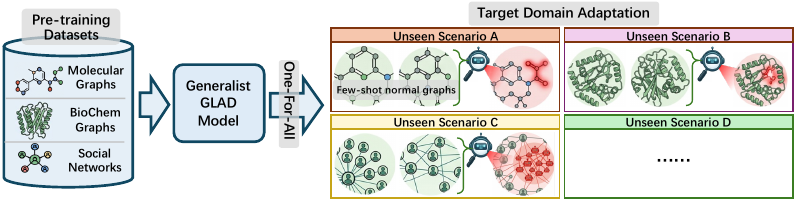}
    \caption{Generalist GLAD paradigm.}
    \label{fig:intro_paradigm}
  \end{subfigure}
  \hfill
  \begin{subfigure}{0.25\linewidth}
    \centering
    \includegraphics[height=2.6cm]{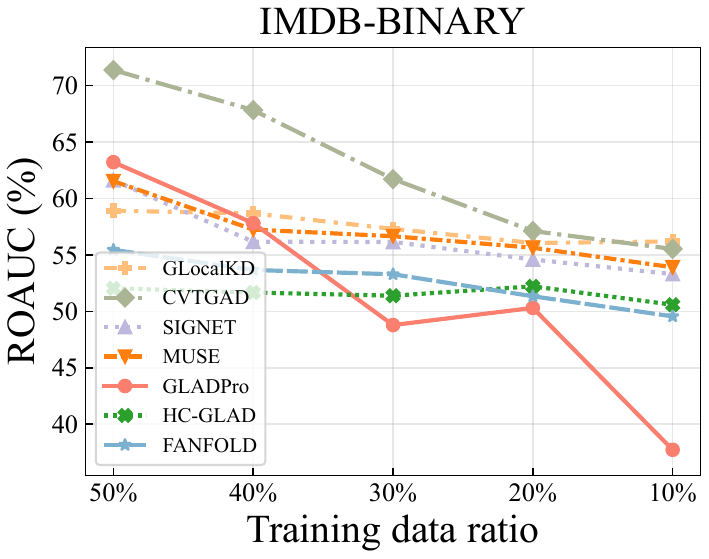}
    \caption{AUROC comparison.}
    \label{fig:intro_prelim}
  \end{subfigure}
  \vspace{-1mm}
  \caption{(a) Sketch map the generalist GLAD paradigm, where one model fits multiple new datasets. (b) Performance of unsupervised GLAD methods reduces with the reduction of training set size.}
  \label{Figure:Intro:Prelim}
  \vspace{-1mm}
\end{figure*}

Given the aforementioned limitations, a natural question arises: \textit{Can we train a ``one-for-all'' GLAD model that can generalize to detecting anomalous graphs across diverse application scenarios, without heavy adaptation cost?} Following the previous success of generalist anomaly detection methods~\cite{liu2024arc,zhu2024toward}, developing a \textbf{generalist GLAD} approach is a promising solution to the question. As shown in Figure~\ref{fig:intro_paradigm}, a generalist GLAD model can be trained on a large collection of diverse datasets, acquiring strong generalization capabilities for various GLAD tasks. When applied to an arbitrary unseen target dataset, it only requires a few-shot normal samples for minimal-cost adaptation to detect anomalous graphs. 
As the knowledge learned from large-scale pre-training can be transferred to target datasets, the generalist GLAD paradigm eliminates the need for heavy dataset-specific training, reducing both data requirements and training costs.

Despite the promising potential of the generalist paradigm, it is non-trivial to develop a generalist GLAD approach. 
While recent efforts such as ARC~\cite{liu2024arc} and AnomalyGFM~\cite{qiao2025anomalygfm} have promising generalist capabilities for node-level graph anomaly detection, extending such paradigms to GLAD comes with unique challenges. 
Firstly, \textbf{(C1) How to learn generalizable patterns from diverse training datasets?} Different graph-level applications lie in different semantic spaces and different structural priors. As a result, the characteristics of normal and abnormal graphs can vary widely across domains, which makes obtaining generalizable knowledge difficult. For example, in a social network, a fraudster community is usually characterized by dense local structures due to spamming behavior, while in molecular graphs, a small motif that exhibits specific functional groups can determine biotoxicity. In this case, how to develop a generalizable model that can capture transferable patterns across diverse domains remains a fundamental challenge. 
Moreover, \textbf{(C2) How can models effectively adapt to a new test domain under limited supervision?} 
Even if a generalist model can learn transferable patterns across domains, making reliable prediction on a target dataset still requires leveraging a few-shot set of normal samples, which characterize the specific distributional properties of the dataset. Nevertheless, unsupervised GLAD models typically rely heavily on sufficient data to capture the normal distribution: as shown in Figure~\ref{fig:intro_prelim}, their performance degrades sharply as the amount of training data decreases. In this context, how to effectively exploit the limited few-shot normal samples for efficient adaptation of pre-trained generalist GLAD models to unseen domains still remains an open problem.

To address the aforementioned challenges, we propose \ourmethod, a \textbf{Gen}eralist \textbf{GLAD} approach based on discriminative subgraph extraction. Our core idea is to learn transferable \textit{anomaly-discriminative subgraphs} (e.g., the substructures in red in Figure~\ref{fig:intro_paradigm}) through cross-domain pre-training, and adapt them to unseen domains via discrepancy-based evidence modeling under few-shot supervision. 
Specifically, to tackle \textbf{(C1)}, \ourmethod incorporates a memory-guided discriminative subgraph extraction model as its backbone network, learning to capture shared fundamental knowledge for identifying the most anomaly-discriminative substructures that distinguish normal and anomalous graphs from arbitrary graph samples. 
To enhance transferable knowledge learning, \ourmethod empowers cross-domain pre-training by storing learned knowledge in a memory bank, and also facilitating task-relevant knowledge retrieval with the help of contextual priors. 
Moreover, to address \textbf{(C2)}, we construct multi-dimensional discrepancy scores between the discriminative subgraph of each graph sample and the contextual information, using them as prediction evidence. In this way, \ourmethod not only leverages few-shot contextual samples for domain-specific adaptation, but also captures anomaly evidence from fine-grained instance-level cues to coarse-grained cross-domain knowledge. With the discrepancy-based evidence, we can perform efficient adaptation by learning a lightweight scoring module with a linear layer to support anomaly prediction on new datasets. 
To summarize, this paper makes the following contributions:

\vspace{-1mm}
\begin{itemize}[noitemsep,leftmargin=*,topsep=1.5pt]
    \item \textbf{Problem.}  To the best of our knowledge, we are the first to formalize and investigate the generalist GLAD problem, aiming to detect anomalous graphs from various domains with a single pre-trained GLAD model with limited in-domain supervision when adapting to arbitrary unseen datasets.
    \item \textbf{Method.} We propose a novel generalist GLAD method, \ourmethod, which learns transferable anomaly-discriminative substructures through memory-guided cross-domain pre-training and performs efficient few-shot adaptation via discrepancy-based evidence modeling.
    \item \textbf{Experiments.} We conduct extensive experiments to demonstrate the strong performance of \ourmethod, and illustrate its generalizability and efficiency across 10 benchmark datasets from various application domains.
\end{itemize}

\vspace{-1mm}




\section{Related Work}
\vspace{-1mm}

\noindent\textbf{Graph-Level Anomaly Detection} aims to identify anomalous graphs that significantly deviate from the majority of normal graphs in a dataset~\cite{ma2021comprehensive, wang2025unifying}. Existing methods can be broadly categorized into two groups, namely, reconstruction-based methods and contrastive learning-based methods. The former aim to reconstruct input graphs and detect anomalies based on reconstruction errors, assuming that anomalous graphs are harder to reconstruct~\cite{luo2022deep, cao2024fanfold, kim2024rethinking, zhao2026fedcigar}. The latter learn graph representations by contrasting multiple views of the data, using consistency or agreement between views to distinguish normal patterns from anomalies~\cite{ li2023cvtgad, liu2023towards, yang2025global, fu2026hc}. There are also several studies that leverage advanced techniques, such as evolutionary mapping~\cite{ma2023towards}, mixture-of-experts~\cite{cai2025moegad}, and knowledge distillation~\cite{ma2022deep}, to address the GLAD problem. Existing GLAD methods usually follow a one-dataset-one-model paradigm, where models are trained using either purely normal data (one-class unsupervised setting~\cite{wang2025unifying}) or labeled normal and anomalous data (supervised setting~\cite{zhang2022dual, ma2023towards, DongRQGNN}), and evaluated on the same dataset. This paradigm increases the requirement for large amounts of labeled or curated data and limits its broad applicability.

\noindent\textbf{Generalist Anomaly Detection} is an emerging paradigm that aims to train a universal anomaly detection model on diverse datasets and generalize it to unseen datasets with minimal supervision and cost~\cite{zhu2024toward, pan2026correcting}. Recently, generalist anomaly detection approaches have achieved significant progress and demonstrated strong performance across various domains, such as industrial image data~\cite{zhu2024toward}, time-series data~\cite{shentu2025towards}, relational data~\cite{li2026relationalAD}, and tabular data~\cite{li2026towards}. In text data, such techniques can be utilized to spot hallucination~\cite{qian2026dynhd, chen2025multi} or fraudsters~\cite{pan2026camera}. Recent visual approaches further enhance generalization through residual feature modeling~\cite{yao2024resad}, normal-abnormal guidance~\cite{wang2026normal}, and residual-guided normality learning~\cite{wang2026hunting}. For graph data, recent studies investigate generalist approaches for node-level anomaly detection~\cite{liu2024arc, qiao2025anomalygfm, liu2026rethinking}. While these methods have shown promising performance, they cannot be directly applied to GLAD due to fundamental differences between node-level and graph-level abnormal patterns: node-level methods typically rely on several assumptions (e.g., the homophily assumption~\cite{pan2025label}), which may not hold in the graph-level counterpart~\cite{PanSurvey}.


A more comprehensive and detailed literature review is given in Appendix~\ref{Appendix:RelatedWork}.

\section{Problem Statement}
\noindent\textbf{Notations.}
In general, a graph can be presented as $\mathcal{G}=(\mathcal{V},\mathcal{E})$, where $\mathcal{V}=\{v_{1},\dots,v_{n}\}$ is the set of $n$ nodes and $\mathcal{E}\subseteq\mathcal{V}\times\mathcal{V}$ is the set of edges, which can also be presented in adjacency matrix form $\mathbf{A}\in\{0,1\}^{n\times n}$, with $\mathbf{A}_{i, j}=1$ iff $(v_{i},v_{j})\in\mathcal{E}$. For a node $v\in\mathcal{V}$, $\mathcal{N}(v)$ denotes its one-hop neighborhood and $\deg(v)=|\mathcal{N}(v)|$ its degree. 

\noindent\textbf{Conventional GLAD Problem.} 
GLAD aims to identify anomalous graphs from a collection of graphs by learning an anomaly scoring function $f:\mathcal{G} \xrightarrow{} \mathbb{R} $ where higher scores correspond to greater degrees of abnormality. The function $f$ is optimized on the training split of target datasets $D=\{\mathcal{G}_1,\dots, \mathcal{G}_K\}$. The mainstream studies of GLAD usually follow an one-class (sometimes a.k.a. unsupervised) setting, 
where the model is trained on a collection of normal graphs and is then applied to detect anomalies in previously unseen graphs. 
The inductive property makes GLAD challenging, as anomalous patterns are not observed during training. 

\noindent\textbf{Generalist GLAD Problem.}
In this paper, we study the \emph{generalist GLAD} problem, which aims to develop a generalist GLAD model capable of detecting anomalous graphs across diverse application scenarios, with only lightweight fine-tuning on few-shot normal samples from target dataset. 
Formally, we define $\mathcal{T}^{\text{train}} = \{\mathcal{D}^{\text{train}}_1, \dots, \mathcal{D}^{\text{train}}_N\}$ as a collection of training datasets, where each dataset $\mathcal{D}^{\text{train}}_i = \{G^{\text{train}}_{i,j}\}_{j=1}^{N_i}$ consists of $N_i$ normal graphs from a particular application scenario. The goal is to train a generalist GLAD model $f$ on $\mathcal{T}^{\text{train}}$ such that it can effectively detect anomalies in unseen test datasets $\mathcal{T}^{\text{test}}$, where $\mathcal{D}^{\text{test}}_i \in \mathcal{T}^{\text{test}}$.  During inference, we adopt the commonly used \emph{normal few-shot} setting~\cite{liu2024arc,qiao2025anomalygfm,zhang2026ta, zhu2024toward}. Specifically, for each test dataset $\mathcal{D}^{\text{test}}_i$, only a small number ($n_k$) of normal graphs are available for lightweight adaptation. The model $f$ is then required to assign anomaly scores to the remaining graphs in the dataset. In this paper, we assume that all graphs are unattributed to cover a wide range of real-world scenarios, because \ding{182}~data from several domains do not come with node or edge features~\cite{wang2025unifying}, and \ding{183}~node features may play a less critical role in GLAD~\cite{xiao2024motif, cai2024towards}.

\section{Method}
\begin{figure*}
    \vspace{-2mm}
  \centering
   \includegraphics[width=\linewidth]{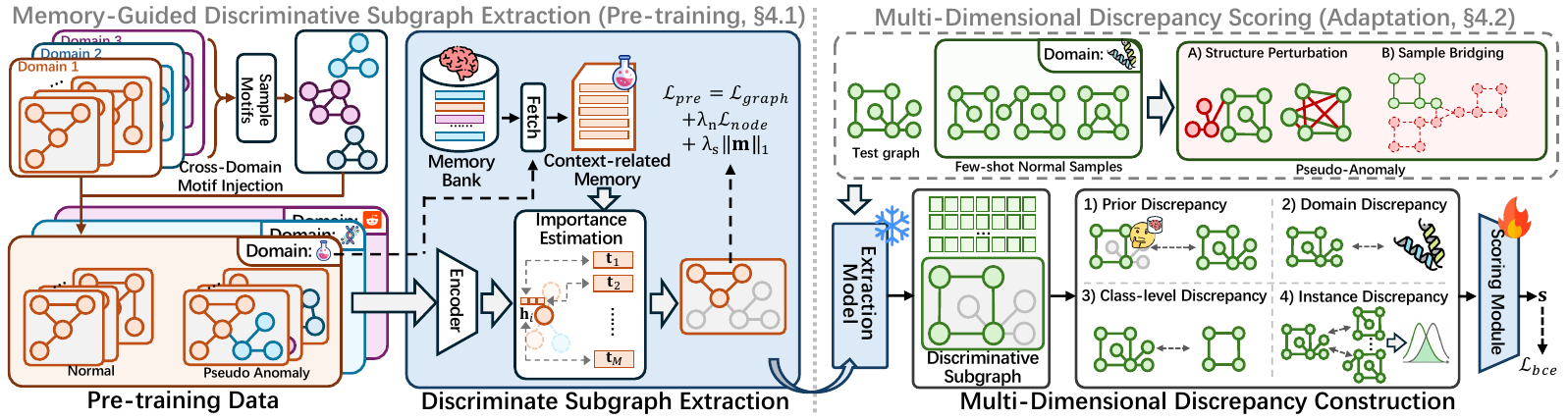}
   \vspace{-2mm}
  \caption{The overall framework of \ourmethod. During the pre-training phase, a discriminative subgraph extraction model learns the transferable patterns for generalist GLAD. During adaptation, multi-dimensional discrepancy is constructed for anomaly scoring on target dataset. Code available at https://github.com/CampanulaBells/Generalist-Graph-Level-Anomaly-Detection. }\label{Figure:Method:Architecture}
  \vspace{-2mm}
\end{figure*}

%

In this section, we introduce~\ourmethod, a generalist GLAD approach that can identify anomalous graph samples on arbitrary testing datasets with few-shot adaptation. As shown in Figure~\ref{Figure:Method:Architecture}, the pre-trained model in \ourmethod learns to \textit{extract discriminative subgraphs} for anomaly discrimination, with a \textit{memory bank} to capture transferable knowledge and a \textit{cross-domain motif injection strategy} to generate pseudo-anomalies (Sec.~\ref{subsec:pre-train}). During adaptation phase, \ourmethod leverages few-shot normal samples to compute multi-dimensional discrepancy for anomaly detection, with a lightweight scoring module for final prediction (Sec.~\ref{subsec:adaptation}).

\subsection{Memory-Guided Discriminative Subgraph Extraction}\label{subsec:pre-train}

To establish a generalist model, a fundamental challenge is to capture generalizable patterns across diverse graph domains for normal-anomaly discrimination. However, such patterns can vary significantly across domains due to differences in graph structures and distributions, making it difficult to identify a specific type of pattern as transferable knowledge. To address this challenge, instead of relying on a specific type of pattern, we propose a generalizable graph neural network (GNN) model that learns to \textbf{extract the most discriminative substructures} from arbitrary graph samples across diverse domains, serving as the transferable backbone of \ourmethod~(Sec.~\ref{subsubsec:encoder}). The discriminative subgraph can serve as key clues for distinguishing normal and abnormal samples, facilitating downstream discrimination. To capture domain-specific knowledge during subgraph extraction, we introduce a memory bank to store transferable patterns, supporting subgraph extraction with domain-relevant information~(Sec.~\ref{subsubsec:memory}). To optimize the model using normal-only data, we propose a cross-domain motif injection strategy to generate pseudo-anomalous samples for training, together with a tailored objective function to guide the learning process~(Sec.~\ref{subsubsec:train}).

 

\subsubsection{GNN-based Subgraph Extractor}\label{subsubsec:encoder}

While anomaly patterns across different domains can be diverse, the abnormality of anomalous graphs is usually characterized by a few key substructures~\cite{liu2023towards, yang2025global}. For example, in molecular graphs, abnormality (e.g., chemical instability) can often be attributed to specific functional groups, while in social networks, anomalous behavior may be associated with unusual local irregular interaction motifs. Motivated by this, we propose a GNN backbone model that identifies the most discriminative subgraphs from various graph samples, learning the transferable patterns for GLAD across diverse domains. This capability is generalizable to unseen graphs for anomaly discrimination~\cite{ma2023towards}.


\noindent\textbf{Structural Attribute Construction.} 
As we focus on a more general unattributed setting, the first step is to construct a unified and semantically rich node attributes as the input features for the GNN. 
To this end, we incorporate structural features as node attributes. Specifically, for each graph $\mathcal{G}$, node attributes are constructed by concatenating one-hot encodings of node degrees with random-walk diffusion-based structural encodings~\cite{dwivedi2022graph}, i.e, 
$\mathbf{X}=\mathbf{X}^{\text{deg}} \oplus \mathbf{X}^{\text{RW}}$. 
This design not only unifies the dimensionality and semantics of node attributes across datasets but also equips them with informative structural signals, thereby providing a consistent and effective foundation for generalist GLAD.

\noindent\textbf{Subgraph Extraction Network.} \label{Method:Pretrain:subsubsec:archi}
Given an input graph $\mathcal{G}=(\mathcal{V}, \mathcal{E})$ with unified node attributes $\textbf{X}$, we aim to extract a discriminative subgraph $\mathcal{G}'=(\mathcal{V}', \mathcal{E}')$ that preserves anomaly-discriminative information while filtering out irrelevant structures. To achieve this, we first obtain node representations $\mathbf{H}$ via a graph encoder, and then compute the node importance scores $\mathbf{p}$ through an importance estimation module $f_{\theta}(\cdot)$:
\begin{equation}
    \mathbf{H} = \mathrm{GNN}(\mathbf{X}, \mathbf{A}),
    \quad \mathbf{p}= f_{\theta}(\mathbf{H}).
\end{equation}
To extract the task-relevant subgraph while obtaining a differentiable node selection, we then apply Gumbel-Softmax with temperature $\tau$ to transform the importance scores into a binary mask:
\begin{equation}
    \mathbf{m} = \mathrm{GumbelSoftmax}(\mathbf{p}; \tau), \quad \mathbf{m} \in \{0,1\}^{|\mathcal{V}|}.
\end{equation}
The resulting mask is used to filter nodes and edges, yielding the extracted subgraph $\mathcal{G}'$. This subgraph focuses the model on the most informative regions of the input graph while ignoring the irrelevant structures, serving as the discriminative evidence for downstream detection.


\subsubsection{Importance Estimation via Memory Retrieval}\label{subsubsec:memory}

In the subgraph extraction network, it is important to ensure that $f_{\theta}(\cdot)$ consistently identifies the most discriminative subgraphs. 
However, what constitutes such discriminative structures is inherently domain-dependent. Due to the diversity of application scenarios, topological patterns may vary significantly across domains, implying that the estimation module $f_{\theta}(\cdot)$ must incorporate domain-specific knowledge. To improve the domain adaptability of $f_{\theta}(\cdot)$, we equip it with a memory bank that encodes anomaly-related knowledge accumulated from diverse training domains. Given a new dataset, we can retrieve domain-relevant memory conditioned on its target context, thereby guiding the extraction process.

 


Specifically, we maintain a set of $M$ trainable memory tokens $\mathbf{T} = \{\mathbf{t}_1, \ldots, \mathbf{t}_M\}$, where each token is responsible for identifying a discriminating graph pattern associated with anomaly detection across datasets. Then, given a target dataset $\mathcal{D}$ with available graph samples $\mathcal{S}\subset \mathcal{D}_i$, we summarize a compact context vector that serves as the condition for retrieval: $\mathbf{c} = \frac{1}{\|\mathcal{S}\|}\sum_{\mathcal{G} \in \mathcal{S}}\mathrm{Readout}_{\mathcal{G}}(\mathbf{X})$, with mean pooling operation for readout. Here, $\mathcal{S}$ can be a batch of normal sample during training, or the few-shot supervision during the adaptation stage. 
The obtained context vector $\mathbf{c}$ is then used to adaptively retrieve relevant knowledge from the memory bank. In detail, we choose the tokens with top-$K$ relevance by computing cosine similarity between $\mathbf{c}$ and memory tokens, resulting in the retrieved token set $\tilde{\mathbf{T}}$ which encodes the domain-specific knowledge. Finally, the node importance is estimated by measuring the alignment between node embeddings and the retrieved memory tokens:
\begin{equation}
    \mathbf{p}_i = \max_{\mathbf{t} \in \tilde{\mathbf{T}}} \ \mathrm{cos}(\mathbf{h}_u, \mathbf{t}),
\end{equation}
where $\mathbf{h}_u$ denotes the encoded embedding of node $u$ in $\mathcal{G}$, and $\mathrm{cos}(\cdot, \cdot)$ is the cosine similarity. By obtaining the maximum similarity score, we capture the most significant aligned pattern, ensuring the extracted subgraph is discriminative. In this way, the memory bank supplies transferable knowledge, while context-conditioned retrieval ensures domain-specific subgraph extraction. 

\subsubsection{Model Training}\label{subsubsec:train}

To train a generalist model for GLAD, a straightforward but  effective learning objective is teaching it to distinguish between anomalous and normal samples. 
However, annotating abnormal graph samples demands heavy domain expertise, 
making the training data for GLAD usually under a one-class assumption, i.e., only normal samples are available~\cite{liu2023towards}. To optimize the generalist model in such an unsupervised pre-training setting, we generate pseudo-anomalies via a \textit{cross-domain motif injection} strategy, and leverage them to guide model training with a bi-level loss function.

\noindent\textbf{Cross-Domain Motif Injection.} 
To generate pseudo-anomalies for training datasets from diverse domains, we propose a cross-domain motif injection strategy that leverages motifs from one domain to perturb graphs in another. The detailed procedure can be found at Appendix~\ref{Appendix:Pretraining}. 
Concretely, given a batch of graphs sampled from a training dataset $\mathcal{D}^{\text{train}}_i \in \mathcal{T}^{\text{train}}$, we treat these graphs as normal samples for the corresponding graph application, allocated with pseudo label $\tilde{y}=0$. Pseudo-anomalies ($\tilde{y}=1$) are then constructed by incorporating cross-dataset substructure, where each normal graph is augmented with a motif randomly sampled from graphs drawn from other datasets. As such subgraphs originate from different data distributions, they can disrupt the inherent structural patterns of the original graph, naturally introducing anomalous signals. To merge the graph and the motif, we link them with additional edges to form a single connected graph. To further provide fine-grained supervision signals, we assign a node-level pseudo-label vector $\tilde{\mathbf{y}}^{(n)} \in \{0,1\}^{|\tilde{\mathcal{V}}_b|}$ for each graph, where each entry indicates whether a node is injected from a motif: $\tilde{\mathbf{y}}^{(n)}_u = 1$ if node $u$ belongs to any injected motif, and $\tilde{\mathbf{y}}^{(n)}_u = 0$ otherwise. 


\noindent\textbf{Bi-Level Training Loss.} 
To ensure the model can distinguish anomalies across diverse scenarios, we train the model by a pseudo-label-guided bi-level loss. 
We first formulate a \textbf{graph-level objective} that encourages the extracted subgraph $\mathcal{G}'$ to be semantically consistent with domain-specific knowledge while deviating from anomalous ones. Given that the memory tokens encode prototypical discriminative patterns, we compute the similarity between the graph-level representation $\mathbf{z}_{\mathcal{G}'}=\mathrm{Readout}_{\mathcal{G}'}(\mathbf{H})$ and the aggregated in-domain memory token $\mathbf{t}_\mathbf{c}= \mathrm{Mean}(\mathbf{T}')$, and optimize it with a binary cross-entropy (BCE) objective:
\begin{equation}
\mathcal{L}_{\mathrm{graph}}
= \mathrm{BCE}\bigl(\sigma(1-\mathrm{cos}(\mathbf{z}_{\mathcal{G}'}, \mathbf{t}_\mathbf{c})),\, \tilde{y}\bigr),
\end{equation}
where $\mathrm{cos}(\cdot,\cdot)$ denotes cosine similarity and $\sigma(\cdot)$ is a sigmoid function. This objective encourages the extracted in-domain knowledge to be aligned with normal graphs while diverging from anomalies, thereby promoting cross-domain generalization.

To provide node-level supervisions, we leverage the injected motifs to form \textbf{node-level objective}. Specifically, we utilize the labeled injected nodes as hard negatives and penalize false positive predictions, encouraging the model to exclude them from the learned discriminative subgraph:
\begin{equation} 
\mathcal{L}_{\text{node}} = - \mathbb{E}_{i \sim \tilde{\mathbf{y}}^{(n)}_i=1}\!\left[\log(1 - m_i)\right]. 
\end{equation}

Finally, we integrate the graph- and node-level objectives into a unified pre-training loss: 
\begin{equation}
\mathcal{L}_{\mathrm{pre}}
= \mathcal{L}_{\mathrm{graph}}
+ \lambda_{\mathrm{n}} \mathcal{L}_{\mathrm{node}}
+ \lambda_{\mathrm{s}} \|\mathbf{m}\|_1,
\end{equation}
where the sparsity regularization term $\|\mathbf{m}\|_1$ encourages the model to focus on compact and informative substructures, while $\lambda_{\mathrm{n}}$ and $\lambda_{\mathrm{s}}$ are balance hyperparameters. Overall, the training objective utilizes the informative pseudo-labels to guide the learning of \ourmethod with both coarse-grained and fine-grained losses, grounding robust generalization across diverse graph application domains.

\subsection{Multi-Dimensional Discrepancy Scoring for Few-Shot Adaptation}\label{subsec:adaptation}
Although the pre-trained backbone model has the capability to capture informative subgraphs that are useful for anomaly discrimination, simply applying it to unseen datasets may not always generalize well due to domain shift and variations in abnormality patterns. To achieve effective target-dataset adaptation, a key idea is to fully exploit the knowledge encoded in the pre-trained model and the memory bank, together with the few-shot samples that reflects the characteristics of the target domain. To this end, we construct multi-dimensional discrepancy signals that characterize the deviation of the discriminative subgraph of an incoming sample from multiple perspectives, including prior knowledge, domain-level context, and feature semantics. To adaptively leverage these discrepancy signals for anomaly detection, we further learn a lightweight scoring module with the few-shot normal samples, ensuring reliable predictions are made based on the discrepancy-based evidence. 



\noindent\textbf{Multi-Dimensional Discrepancy Construction.} 
\label{Method:Finetune:Multi-Dimensional Discrepancy Construction}
While the extracted subgraph captures key characteristics of each graph sample, it is not straightforward to characterize its position within the normal distribution of the target dataset for anomaly detection.
To achieve this, we model anomaly evidence as discrepancy scores across multiple semantic spaces. We construct a set of complementary discrepancy signals for each incoming graph, providing a holistic characterization of it. 
Specifically, we construct multi-dimensional discrepancy measures from four perspectives. 

\vspace{-1mm}
\begin{enumerate}[leftmargin=*, itemsep=0pt, topsep=0pt, parsep=0pt, partopsep=0pt]
    \item \textbf{Prior Discrepancy}, which measures the deviation of the discriminative subgraph embedding $\mathbf{z}_{\mathcal{G}'}$ from the pre-trained semantic prior, characterized by its similarity to the pre-trained prototype $\mathbf{t_{\mathbf{c}}}$. This perspective captures how well the current graph aligns with the general knowledge learned during pre-training, serving as a basic indicator for anomaly detection. 
    \item \textbf{Domain Discrepancy}, which measures the deviation of $\mathbf{z}_{\mathcal{G}'}$ from the domain-level context, characterized by its distance to the global domain prototype $\mathbf{c}$. This perspective reflects the extent to which the current graph deviates from the overall data distribution, providing a coarse-grained signal of distributional shift and potential anomaly.
    \item \textbf{Class-level Discrepancy}, which evaluates the mismatch between $\mathbf{z}_{\mathcal{G}'}$ and the class prototype $\mathbf{q}$ calculated as the average representations of few-shot normal samples. This perspective characterizes how well the current graph adheres to the semantic pattern of the normal class, offering a more targeted cue for distinguishing abnormal instances.
    \item \textbf{Instance-level Discrepancy}, which captures the inconsistency between the similarity pattern of the query graph to support instances and the intrinsic relational structure within the support set. This perspective reveals whether the current graph maintains the fine-grained relational regularities of normal data, yielding a distribution-aware signal beyond point-wise comparisons.
\end{enumerate}
\vspace{-1mm}

Notably, the choice of similarity/distance metrics (e.g., cosine and L2) can influence the characterization of discrepancies; therefore, we adopt multiple metrics for certain discrepancies to improve coverage. A formal formulation of the discrepancy scores is provided in Appendix~\ref{Appendix:discrepancy}.


\noindent\textbf{Lightweight Scoring Module.} 
In practice, different datasets may exhibit varying dependence on different discrepancy scores for anomaly detection, due to differences in data distributions. To account for this, we learn a lightweight scoring module, which adopts a linear combination of discrepancy scores: $s = \sigma(\mathbf{w}^\top \mathbf{s})$, where $s$ is the anomaly score of a query graph, $\mathbf{s}$ is the vector composed of multiple discrepancy scores, and $\sigma$ is a Sigmoid function. $\mathbf{w} \in \mathbb{R}^{k \times 1}$ is the only parameter that is learnable during the adaptation phase, while the rest of the model remains frozen. 

\vspace{-1mm}

To optimize $\mathbf{w}$, a straightforward approach is to use a binary cross-entropy loss with binary normal/anomalous labeled samples. Nevertheless, the few-shot set only contains normal instances, leading to an urgent need for a mechanism to distinguish pseudo-anomalies from normal samples. As large-scale training datasets are not available at this stage, we employ two simple yet effective strategies to distinguish pseudo-anomalies from few-shot normal samples (details of pseudo-anomaly construction is given in Appendix~\ref{Appendix:Finetuning}): \ding{182}~\textbf{Structure perturbation}, which constructs outlier graphs by randomly shuffling graph structures to destroy meaningful relational patterns, and \ding{183}~\textbf{Sample bridging}, which generates outlier graphs by merging normal graphs to simulate over-amplified in-domain structural patterns. With the original samples as normal instances and pseudo-anomalies as negative instances, we can optimize $\mathbf{w}$ with only a few gradient steps during adaptation phase. Algorithm description and complexity analysis of \ourmethod are in Appendices~\ref{Appendix:MethodAlgorithm} and \ref{Appendix:Complexcity}. 

\vspace{-2mm}

\section{Experiments}
\vspace{-1mm}
\subsection{Experimental Setup}
\vspace{-1mm}
\noindent \textbf{Datasets. }
We train both the baseline approaches and \ourmethod on one set of graph datasets and evaluate them on a separate set of datasets. For a comprehensive evaluation, we utilize real-world GLAD datasets drawn from diverse domains, including molecular graphs, biochemical graphs, and social networks. Following~\cite{liu2024arc}, we train the models on the largest dataset within each domain and evaluate them on the remaining datasets. Specifically, the training datasets include NCI1, DD, and COLLAB, while the testing datasets consist of molecular graphs: AIDS, BZR, COX2, DHFR, MUTAG, PTC\_MR; biochemical graphs: ENZYMES, PROTEINS; and social networks: IMDB-BINARY, and REDDIT-BINARY. We also denote the collection of molecular and chemical graphs as science graphs. For more detailed information, please refer to the Appendix~\ref{Appendix:Setup}. 

\vspace{-1mm}

\noindent \textbf{Baselines.} We compare \ourmethod with several unsupervised GLAD baselines, including contrastive methods such as GLocalKD~\cite{ma2022deep}, GLADPro~\cite{yang2025global}, CVTGAD~\cite{li2023cvtgad},  SIGNET~\cite{liu2023towards}, and  HC-GLAD~\cite{fu2026hc}; reconstruction-based methods such as FANFOLD~\cite{cao2024fanfold} and MUSE~\cite{kim2024rethinking}. 

\noindent \textbf{Evaluation and Implementation. }
Following~\cite{wang2025unifying}, we adopt AUROC and AUPRC as evaluation metrics for GLAD. We report the mean AUROC/AUPRC along with standard deviations over five independent runs. We train \ourmethod jointly on all datasets in $\mathcal{T}^\text{train}$ and evaluate it on each dataset in $\mathcal{T}^\text{test}$ with few-shot tuning, where $n=10$ normal graphs are provided as supervision by default. For all baselines, we consider two settings: "Pre-training+Few-shot Fine-tuning" and "Few-shot Fine-tuning Only".  In the latter, baselines are trained directly using only the few-shot normal supervision to avoid potential performance degradation caused by negative transfer. We utilize random search to determine the optimal hyperparameters for both the baselines and \ourmethod (Details in Appendix~\ref{Appendix:Setup}). Since our goal is to train generalist GLAD models, we do not perform dataset-specific hyperparameter tuning, but instead use the same set of hyperparameters for all testing datasets. 

\begin{table*}[t]
\vspace{-3mm}
\caption{Anomaly detection performance in terms of AUROC (in percent, mean$\pm$std). Highlighted are the results ranked \textcolor{firstcolor}{\textbf{\underline{first}}}, \textcolor{secondcolor}{\underline{second}}, and \textcolor{thirdcolor}{\underline{third}}. OOM indicates out-of-memory.}
\centering
\label{tab:results_main_auc}
\resizebox{1\textwidth}{!}{
\begin{tabular}{l|cccccccccc}
\toprule
Method & AIDS & BZR & COX2 & DHFR & MUTAG & PTC\_MR & ENZYMES & PROTEINS & IMDB-B & REDDIT-B\\
\midrule

\rowcolor{Gray}
\multicolumn{11}{c}{Pre-training+Few-shot Fine-tuning} \\

  GLocalKD & $50.67{\scriptstyle\pm0.12}$ & $50.29{\scriptstyle\pm0.04}$ & \third{$52.73{\scriptstyle\pm0.07}$} & $56.90{\scriptstyle\pm0.11}$ & $79.55{\scriptstyle\pm0.06}$ & $49.34{\scriptstyle\pm0.03}$ & $50.12{\scriptstyle\pm0.06}$ & $49.90{\scriptstyle\pm0.06}$ & \third{$67.82{\scriptstyle\pm0.23}$} & $67.82{\scriptstyle\pm0.23}$\\
  CVTGAD & $40.65{\scriptstyle\pm0.00}$ & $45.96{\scriptstyle\pm0.01}$ & $52.00{\scriptstyle\pm0.03}$ & $55.81{\scriptstyle\pm0.00}$ & \second{$89.55{\scriptstyle\pm0.01}$} & $51.83{\scriptstyle\pm0.01}$ & $45.31{\scriptstyle\pm0.00}$ & $40.36{\scriptstyle\pm0.71}$ & OOM & OOM\\
  SIGNET & \second{$90.59{\scriptstyle\pm0.00}$} & $21.46{\scriptstyle\pm0.02}$ & \second{$53.56{\scriptstyle\pm0.01}$} & $53.56{\scriptstyle\pm0.01}$ & $82.47{\scriptstyle\pm0.02}$ & $53.19{\scriptstyle\pm0.02}$ & $47.28{\scriptstyle\pm0.00}$ & $57.19{\scriptstyle\pm0.01}$ & $50.66{\scriptstyle\pm0.00}$ & $50.66{\scriptstyle\pm0.00}$\\
  MUSE & $77.30{\scriptstyle\pm1.73}$ & $52.77{\scriptstyle\pm0.42}$ & $45.27{\scriptstyle\pm0.27}$ & $57.24{\scriptstyle\pm0.88}$ & $84.05{\scriptstyle\pm1.67}$ & $47.32{\scriptstyle\pm0.33}$ & $52.48{\scriptstyle\pm0.47}$ & $52.57{\scriptstyle\pm1.25}$ & \second{$67.84{\scriptstyle\pm1.06}$} & $67.84{\scriptstyle\pm1.06}$\\
  GLADPro & $50.00{\scriptstyle\pm0.00}$ & $47.39{\scriptstyle\pm1.72}$ & $48.10{\scriptstyle\pm0.69}$ & $53.04{\scriptstyle\pm1.84}$ & $50.00{\scriptstyle\pm0.00}$ & \third{$53.33{\scriptstyle\pm3.61}$} & $53.13{\scriptstyle\pm0.63}$ & $38.51{\scriptstyle\pm1.36}$ & $64.16{\scriptstyle\pm7.36}$ & $64.16{\scriptstyle\pm7.36}$\\
  HC-GLAD & OOM & OOM & OOM & OOM & OOM & OOM & OOM & OOM & OOM & OOM\\
  FANFOLD & $73.54{\scriptstyle\pm0.00}$ & $50.36{\scriptstyle\pm0.01}$ & $37.94{\scriptstyle\pm0.02}$ & $58.18{\scriptstyle\pm0.00}$ & $87.89{\scriptstyle\pm0.02}$ & $41.70{\scriptstyle\pm0.01}$ & $49.50{\scriptstyle\pm0.00}$ & $49.41{\scriptstyle\pm0.00}$ & $56.64{\scriptstyle\pm0.00}$ & $18.51{\scriptstyle\pm0.00}$\\
\midrule

\rowcolor{Gray}
\multicolumn{11}{c}{Few-shot Fine-tuning Only} \\

  GLocalKD & $57.71{\scriptstyle\pm2.91}$ & \second{$58.91{\scriptstyle\pm4.54}$} & $52.14{\scriptstyle\pm2.39}$ & \second{$63.34{\scriptstyle\pm1.54}$} & \third{$89.54{\scriptstyle\pm0.23}$} & $48.83{\scriptstyle\pm0.99}$ & $47.60{\scriptstyle\pm2.08}$ & $53.05{\scriptstyle\pm2.89}$ & $59.79{\scriptstyle\pm1.63}$ & \third{$72.37{\scriptstyle\pm1.56}$}\\
  CVTGAD & $69.49{\scriptstyle\pm7.78}$ & $43.30{\scriptstyle\pm3.80}$ & $43.36{\scriptstyle\pm3.96}$ & $46.24{\scriptstyle\pm1.47}$ & $79.41{\scriptstyle\pm8.93}$ & $43.29{\scriptstyle\pm2.79}$ & \second{$54.50{\scriptstyle\pm2.90}$} & \second{$69.81{\scriptstyle\pm0.86}$} & $55.51{\scriptstyle\pm4.79}$ & OOM\\
  SIGNET & $68.21{\scriptstyle\pm5.49}$ & $44.60{\scriptstyle\pm3.52}$ & $44.38{\scriptstyle\pm3.02}$ & $35.21{\scriptstyle\pm3.12}$ & $68.86{\scriptstyle\pm14.45}$ & $50.93{\scriptstyle\pm0.55}$ & $41.28{\scriptstyle\pm2.02}$ & $55.48{\scriptstyle\pm1.78}$ & $45.97{\scriptstyle\pm2.37}$ & $52.36{\scriptstyle\pm1.95}$\\
  MUSE & \third{$84.56{\scriptstyle\pm5.82}$} & $47.32{\scriptstyle\pm0.87}$ & $41.80{\scriptstyle\pm1.35}$ & \third{$61.74{\scriptstyle\pm1.91}$} & $85.77{\scriptstyle\pm6.01}$ & $50.76{\scriptstyle\pm0.47}$ & $49.50{\scriptstyle\pm0.33}$ & \third{$61.81{\scriptstyle\pm0.17}$} & $58.18{\scriptstyle\pm2.38}$ & \first{$73.85{\scriptstyle\pm1.82}$}\\
  GLADPro & $70.21{\scriptstyle\pm13.92}$ & \third{$56.24{\scriptstyle\pm3.37}$} & $43.62{\scriptstyle\pm3.80}$ & $43.03{\scriptstyle\pm3.72}$ & $82.96{\scriptstyle\pm8.10}$ & \second{$55.73{\scriptstyle\pm3.05}$} & $53.98{\scriptstyle\pm1.45}$ & $43.44{\scriptstyle\pm4.09}$ & $54.01{\scriptstyle\pm2.89}$ & $64.29{\scriptstyle\pm6.07}$\\
  HC-GLAD & $3.64{\scriptstyle\pm0.55}$ & $37.50{\scriptstyle\pm3.32}$ & $46.16{\scriptstyle\pm1.56}$ & $54.99{\scriptstyle\pm2.66}$ & $84.69{\scriptstyle\pm2.79}$ & $48.94{\scriptstyle\pm0.70}$ & \first{$63.03{\scriptstyle\pm4.45}$} & $37.85{\scriptstyle\pm3.67}$ & $61.85{\scriptstyle\pm5.11}$ & OOM\\
  FANFOLD & $60.95{\scriptstyle\pm18.89}$ & $44.92{\scriptstyle\pm1.04}$ & $41.80{\scriptstyle\pm3.84}$ & $60.28{\scriptstyle\pm2.35}$ & $74.00{\scriptstyle\pm20.09}$ & $47.04{\scriptstyle\pm3.60}$ & $52.15{\scriptstyle\pm3.95}$ & $58.24{\scriptstyle\pm3.61}$ & $51.54{\scriptstyle\pm3.97}$ & $23.24{\scriptstyle\pm2.92}$\\
\midrule
\rowcolor{Gray}
\multicolumn{11}{c}{Generalist} \\
 \ourmethod & \first{$95.84{\scriptstyle\pm0.07}$} & \first{$63.89{\scriptstyle\pm1.18}$} & \first{$62.92{\scriptstyle\pm0.30}$} & \first{$74.96{\scriptstyle\pm0.42}$} & \first{$90.75{\scriptstyle\pm0.17}$} & \first{$60.58{\scriptstyle\pm0.47}$} & \third{$54.37{\scriptstyle\pm1.74}$} & \first{$73.63{\scriptstyle\pm0.16}$} & \first{$74.11{\scriptstyle\pm0.25}$} & \second{$73.63{\scriptstyle\pm0.20}$}\\
\bottomrule
\end{tabular}
}
\end{table*}

\vspace{-1mm}

\subsection{Experimental Results}

\vspace{-1mm}

\noindent \textbf{Performance Comparison. }




\begin{wrapfigure}{r}{0.34\textwidth}
 \centering
 \vspace{-7mm}

 \includegraphics[width=0.34\textwidth]{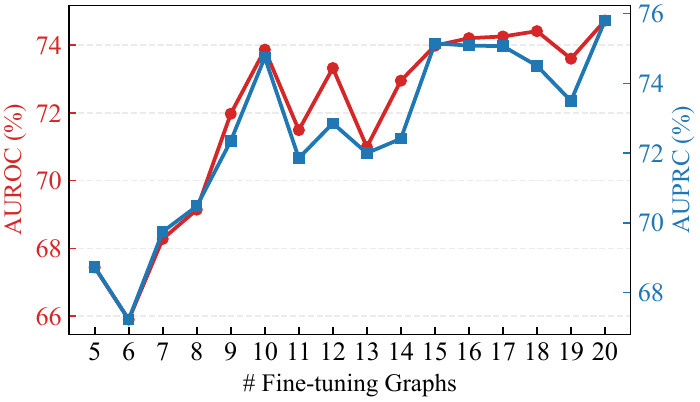}
 \caption{Sensitivity of \#shot.}
 \label{Figure:Expr:Fewshots}

 \vspace{-2mm}
\end{wrapfigure}
Table~\ref{tab:results_main_auc} presents a comparison between \ourmethod and baseline methods in terms of AUROC. Results for AUPRC are provided in Appendix~\ref{appe:exp_main}. We make the following observations: \ding{182} \ourmethod demonstrates strong anomaly detection capability in the generalist GLAD setting, despite using only few-shot fine-tuning on normal samples. Specifically, \ourmethod achieves state-of-the-art performance on 9 out of 10 datasets, while still obtaining competitive performance on the ENZYMES dataset. Moreover, \ourmethod demonstrates significant improvements compared to the best baselines on AIDS, REDDIT, COX2, and DHFR. \ding{183} Pre-training typically does not lead to performance improvements for GLAD baselines, highlighting the limited generalizability of conventional GLAD methods. \ding{184} With a limited amount of training data, baseline methods perform poorly. Notably, in some cases, they achieve an AUROC below 0.5 (worse than random guessing). This highlights that conventional GLAD methods are data-hungry, making them difficult to apply in emerging GLAD scenarios where training data is scarce.

\begin{table*}[t]
\vspace{-2mm}
\caption{Performance (in AUROC) with varying pre-training datasets $\mathcal{T}^\text{train}$. }\label{tab:Ablation_Pretrain_dataset_auc}
\vspace{-1mm}
\centering
\resizebox{1\textwidth}{!}{
\begin{tabular}{l|cccccccccc}
\toprule
Training Datasets & AIDS & BZR & COX2 & DHFR & MUTAG & PTC\_MR & ENZYMES & PROTEINS & IMDB-B & REDDIT-B\\
\midrule
NCI1 & $87.56{\scriptstyle\pm2.68}$ & $56.64{\scriptstyle\pm0.76}$ & $57.72{\scriptstyle\pm1.27}$ & $70.92{\scriptstyle\pm0.60}$ & $75.19{\scriptstyle\pm3.45}$ & $55.83{\scriptstyle\pm1.46}$ & $52.70{\scriptstyle\pm0.45}$ & $63.26{\scriptstyle\pm1.45}$ & $56.14{\scriptstyle\pm1.20}$ & $64.06{\scriptstyle\pm0.16}$\\
DD & $88.33{\scriptstyle\pm4.26}$ & $56.98{\scriptstyle\pm1.25}$ & $57.89{\scriptstyle\pm1.39}$ & $70.87{\scriptstyle\pm0.13}$ & $73.74{\scriptstyle\pm2.93}$ & $55.83{\scriptstyle\pm1.46}$ & $52.82{\scriptstyle\pm0.26}$ & $63.20{\scriptstyle\pm1.40}$ & $55.80{\scriptstyle\pm1.09}$ & $63.46{\scriptstyle\pm1.04}$\\
COLLAB & $87.88{\scriptstyle\pm0.50}$ & $57.48{\scriptstyle\pm1.67}$ & $57.74{\scriptstyle\pm1.76}$ & $71.09{\scriptstyle\pm0.25}$ & $76.25{\scriptstyle\pm1.69}$ & $55.64{\scriptstyle\pm1.43}$ & $52.52{\scriptstyle\pm0.32}$ & $63.30{\scriptstyle\pm1.15}$ & $55.15{\scriptstyle\pm1.38}$ & $65.44{\scriptstyle\pm0.48}$\\
NCI1+DD & $89.72{\scriptstyle\pm2.71}$ & $58.18{\scriptstyle\pm1.52}$ & \third{$58.23{\scriptstyle\pm1.53}$} & $71.10{\scriptstyle\pm0.50}$ & $80.08{\scriptstyle\pm3.99}$ & \second{$57.70{\scriptstyle\pm1.42}$} & $53.19{\scriptstyle\pm0.54}$ & $65.37{\scriptstyle\pm1.09}$ & \third{$66.77{\scriptstyle\pm1.22}$} & \second{$70.96{\scriptstyle\pm1.09}$}\\
NCI1+COLLAB & \third{$91.62{\scriptstyle\pm3.38}$} & \third{$59.92{\scriptstyle\pm1.10}$} & $58.22{\scriptstyle\pm1.46}$ & \third{$72.51{\scriptstyle\pm0.89}$} & \third{$82.55{\scriptstyle\pm2.64}$} & \third{$57.63{\scriptstyle\pm1.40}$} & \third{$53.49{\scriptstyle\pm0.14}$} & \third{$67.04{\scriptstyle\pm0.86}$} & $66.62{\scriptstyle\pm1.76}$ & \third{$69.22{\scriptstyle\pm1.28}$}\\
DD+COLLAB & \second{$91.99{\scriptstyle\pm2.94}$} & \second{$59.99{\scriptstyle\pm1.06}$} & \second{$58.55{\scriptstyle\pm1.32}$} & \second{$72.85{\scriptstyle\pm0.80}$} & \second{$83.09{\scriptstyle\pm2.62}$} & $57.32{\scriptstyle\pm1.21}$ & \second{$53.76{\scriptstyle\pm0.21}$} & \second{$67.33{\scriptstyle\pm1.13}$} & \second{$68.59{\scriptstyle\pm1.44}$} & $65.34{\scriptstyle\pm0.40}$\\
\midrule
All & \first{$95.84{\scriptstyle\pm0.07}$} & \first{$63.89{\scriptstyle\pm1.18}$} & \first{$62.92{\scriptstyle\pm0.30}$} & \first{$74.96{\scriptstyle\pm0.42}$} & \first{$90.75{\scriptstyle\pm0.17}$} & \first{$60.58{\scriptstyle\pm0.47}$} & \first{$54.37{\scriptstyle\pm1.74}$} & \first{$73.63{\scriptstyle\pm0.16}$} & \first{$74.11{\scriptstyle\pm0.25}$} & \first{$73.63{\scriptstyle\pm0.20}$}\\
\bottomrule
\end{tabular}
}
\vspace{-2mm}
\end{table*}

\noindent \textbf{Effectiveness of Data Scale. } To investigate the impact of data scale on the performance of \ourmethod, we vary $\mathcal{T}^\text{train}$. From Table~\ref{tab:Ablation_Pretrain_dataset_auc}, we observe that the performance of \ourmethod consistently improves as more pre-training datasets are incorporated, demonstrating its ability to learn transferable patterns across diverse domains. With the help of memory retrieval, the model can further retrieve domain-relevant knowledge, mitigating the negative transfer issue. Moreover, we report the average performance on social networks (Figure~\ref{Figure:Expr:Fewshots}, more results are in Appendix~\ref{appe:exp_shot}) with varying number of few-shot samples for adaptation. We observe that increasing the number of fine-tuning graphs consistently improves performance, which highlights the effectiveness of our method in leveraging additional adaptation data when available.

\begin{wraptable}{r}{0.4\textwidth}
\centering
\caption{Ablation study.}
\label{tab:ablation_selected}
\resizebox{0.4\textwidth}{!}{
\begin{tabular}{l|ccc}
\toprule
Variant & DHFR & PROTEINS & IMDB-B \\
\midrule
\ourmethod & {74.96} & {73.63} & {74.11} \\
\midrule
\multicolumn{4}{c}{\cellcolor{Gray}{Pre-training}} \\
w/o $\mathcal{L}^\text{node}$ & 70.98 & 66.03 & 58.03 \\
w/o $\|\mathbf{m}\|_1$ & 71.30 & 68.53 & 60.06 \\
\midrule
\multicolumn{4}{c}{\cellcolor{Gray}{Discrepancy Measures}} \\
w/o Prior & 73.05 & 71.66 & 71.15 \\
w/o Domain & 73.77 & 63.76 & 51.65 \\
w/o Cls. & 45.63 & 60.20 & 63.72 \\
w/o Inst. & 74.25 & {73.75} & 72.96 \\
\midrule
\multicolumn{4}{c}{\cellcolor{Gray}{Scoring Module}} \\
MEAN & 28.92 & 41.54 & 42.00 \\
MAX & 40.32 & 61.94 & 30.12 \\
MIN & 29.56 & 38.05 & 43.57 \\
\bottomrule
\end{tabular}
}
\end{wraptable}
\noindent \textbf{Ablation Study.} To verify the effectiveness of each component of \ourmethod, we conduct ablation studies by removing/modifying each design, with the results on selected datasets in Table~\ref{tab:ablation_selected} (full results are in Appendix~\ref{appe:exp_abl}). Specifically, for the pre-training phase, we construct variants by removing $\mathcal{L}\text{node}$ and the $\ell_1$ regularization on $\mathbf{m}$. We find that both components contribute to overall performance. In particular, $\mathcal{L}\text{node}$ has a larger impact, highlighting the importance of informative node-level pseudo anomaly supervision compared to unsupervised regularization. Moreover, we investigate the effectiveness of each discrepancy measure in the fine-tuning stage and the scoring module. Generally, removing any of the discrepancy measures degrades the performance in most cases. Among them, the domain and class discrepancy scores have the most significant impact, as their removal can lead to performance close to or even below random guessing. To assess the necessity of the scoring module, we replace it with simple score aggregation strategies, including mean, max, and min pooling. As illustrated, these replacements significantly degrade performance and can even lead to AUROC scores below 0.5. This suggests that the model fails to distinguish between normal and anomaly-indicative cues even with normal graph supervisions, indicating the importance of our adaptation framework.

\begin{wrapfigure}{r}{0.36\textwidth}
 \centering
 \captionsetup[subfigure]{skip=1pt}
 \vspace{-4mm}

 \begin{subfigure}{\linewidth}
     \centering
     \includegraphics[width=\linewidth]{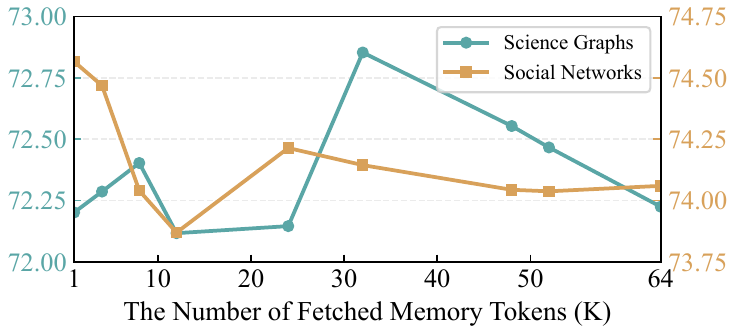}
     \caption{Sensitivity to \# mem. tokens.}
     \label{Figure:Expr:KTokens}
 \end{subfigure}

 \vspace{1mm}

 \begin{subfigure}{\linewidth}
     \centering
     \includegraphics[width=\linewidth]{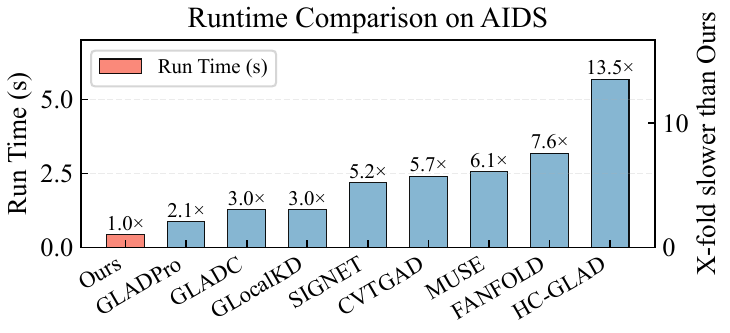}
     \caption{Time comparison.}
     \label{Figure:Expr:Time}
 \end{subfigure}

 \vspace{1mm}

 \begin{subfigure}{\linewidth}
     \centering
     \includegraphics[width=\linewidth]{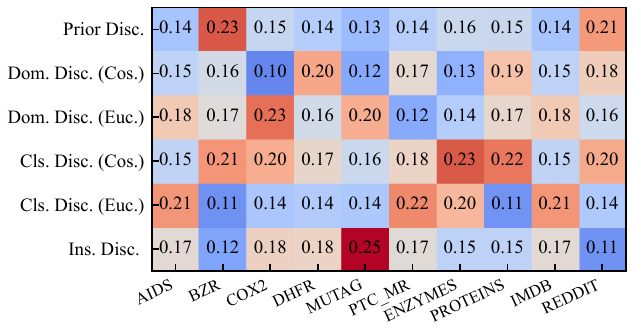}
     \caption{Visualization results.}
     \label{Figure:Expr:MLPActivation}
 \end{subfigure}

 \vspace{-1mm}
 \caption{Analysis results.}
 \label{Figure:Expr:Analysis}
 \vspace{-6mm}
\end{wrapfigure}
\noindent \textbf{Effectiveness of \#Retrieved Tokens.} To verify the effectiveness of retrieved memory tokens, we study the effect of retrieved memory token number $K$ on performance (AUROC) in Figure~\ref{Figure:Expr:KTokens}. Overall, both groups of datasets exhibit a consistent trend: after a certain threshold, increasing $K$ leads to performance degradation, as less relevant tokens may introduce noisy or unhelpful information for the target task. Moreover, scientific graphs and social networks exhibit different sensitivities to $K$, where scientific graphs prefer larger $K$ for diverse motifs, while social networks favor smaller $K$.

\noindent \textbf{Efficiency Analysis. } To investigate the efficiency of \ourmethod, we compare fine-tuning time on the AIDS dataset. As shown in Figure~\ref{Figure:Expr:Time}, \ourmethod outperforms all baselines in terms of efficiency. Compared with our method, the fastest baseline, GLADPro, is still 2.1$\times$ slower. A comprehensive runtime breakdown is further discussed in Appendix ~\ref{appe:efficiency}.

\noindent \textbf{Visualization.} To investigate how the pre-trained \ourmethod adapts to target applications, we visualize the weights of the scoring module. As shown in Figure~\ref{Figure:Expr:MLPActivation}, \ourmethod learns distinct weighting patterns across different test domains. This indicates that the scoring module successfully captures domain-specific dependencies, enabling effective yet efficient few-shot adaptation. 
We also visualized the importance scores of individual memory tokens in Appendix~\ref{appe:exp_vis}.


\section{Conclusion}
In this paper, we take a pioneering step toward addressing the generalist GLAD problem, which aims to detect anomalous graphs across diverse emerging applications under a one-for-all paradigm with only a few-shot normal graph supervision for adaptation. To this end, we propose \ourmethod, a discriminative subgraph extraction-based generalist GLAD approach capable of identifying anomalous graphs with limited in-domain supervision. Extensive experiments on real-world GLAD datasets from various domains demonstrate the effectiveness, generalizability, and efficiency of \ourmethod. The main \textbf{limitation} of \ourmethod is that it still requires few-shot supervision for adaptation and therefore cannot yet achieve fully zero-shot generalist GLAD. \textbf{Future work} may explore incorporating expert knowledge or external knowledge bases to enable zero-shot GLAD across diverse domains.

\begin{ack}
The work of S. Pan was partially supported by the Australian Research Council (ARC) under Grant Nos. DP240101547 and FT210100097. The work of Y. Liu was partially supported by the ARC under Grant No. DE260101172. F. Li was supported by the Australian National Health and Medical Research Council Investigator grants [2041439]. 
\end{ack}

\clearpage

\bibliographystyle{plain}
\bibliography{neurips_2026.bib}

\clearpage
\appendix

\section{Detailed Related Work}
\label{Appendix:RelatedWork}
\subsection{Graph-Level Anomaly Detection}

Graph-level anomaly detection (GLAD) focuses on identifying abnormal graphs that significantly deviate from the dominant graph patterns~\cite{zhang2022dual,niu2023graph,sutherland2003spline}. In many real-world applications, anomaly annotations are scarce or unavailable, making one-class learning the mainstream setting for GLAD. Under this setting, models are trained solely on normal graphs, while anomalous graphs are identified during inference based on their deviations from learned normal patterns (also referred to as the unsupervised setting in prior studies).

Recently, GNN-based approaches have become the prevailing solutions for GLAD and achieved promising performance. 
These methods can be broadly categorized into two groups, namely, reconstruction-based methods and contrastive learning-based methods. The reconstruction-based methods aim to reconstruct input graphs and detect anomalies based on reconstruction errors, assuming that anomalous graphs are harder to reconstruct~\cite{luo2022deep, cao2024fanfold, kim2024rethinking}. For example, GLADC~\cite{luo2022deep} takes autoencoder-based reconstruction as an important supervision signal for GLAD. MUSE~\cite{kim2024rethinking} introduces a two-stage strategy that combines graph reconstruction with reconstruction-error modeling for anomaly identification. FedCIGAR~\cite{zhao2026fedcigar} employs a personalized reconstruction approach for Federated learning~\cite{tan2026influence}.

The contrastive learning-based methods learn graph representations by contrasting multiple views of the data, using consistency or agreement between views to distinguish normal patterns from anomalies~\cite{li2023cvtgad, liu2023towards, yang2025global,fu2026hc}. For instance, SIGNET~\cite{liu2023towards} measures graph abnormality by maximizing the mutual information between the original graph view and its corresponding dual hypergraph representation. CVTGAD~\cite{li2023cvtgad} further introduces a transformer-based cross-view attention mechanism to directly capture the inter-relationship between augmented graph views for anomaly detection.

There are also several studies that leverage advanced techniques, such as evolutionary mapping~\cite{ma2023towards}, mixture-of-experts~\cite{cai2025moegad}, and knowledge distillation~\cite{ma2022deep}, to address the GLAD problem. Specifically, GmapAD~\cite{ma2023towards} adopts an evolutionary search strategy to identify representative nodes and detects anomalies according to the similarity between graphs and the selected representatives. OCGTL~\cite{qiu2022raising} improves graph-level anomaly detection by integrating self-supervised transformation learning to alleviate issues such as hypersphere collapse and performance flip in deep one-class learning. 

Existing GLAD methods usually follow a one-dataset-one-model paradigm, where models are trained using either purely normal data (one-class unsupervised setting~\cite{wang2025unifying}) or labeled normal and anomalous data (supervised setting~\cite{zhang2022dual, ma2023towards, DongRQGNN}), and evaluated on the same dataset. This paradigm increases the requirement for large amounts of labeled or curated data and limits its broad applicability.

\subsection{Generalist Anomaly Detection}

Recently, anomaly detection research has increasingly shifted toward generalist paradigms, where a unified pre-trained model is expected to generalize across different datasets and application scenarios. Instead of retraining separate models for each domain, these methods usually rely on pretrained foundation architectures together with lightweight test-time adaptation strategies such as prompts, contextual examples, or conditional modules.

In computer vision, many generalist approaches are built upon vision-language models (VLMs). For example, WinCLIP reformulates industrial anomaly detection by matching image features with compositional textual prompts at multiple granularities in CLIP, supporting zero-shot and few-shot anomaly recognition without additional task-specific optimization~\cite{chen2023zero,zhouanomalyclip}. Besides VLM-based approaches, generative foundation models have also been explored. Diffusion-based methods reuse denoising processes as generic perceptual priors and identify anomalies according to reconstruction or score discrepancies~\cite{abdi2025zero}. To further reduce dependence on manually designed prompts, InCTRL introduces an in-context residual learning framework, where query samples are compared with a few normal examples during inference for adaptive anomaly estimation~\cite{zhu2024toward}.  Recent works further enhance generalization through residual feature modeling~\cite{yao2024resad}, normal-abnormal guidance~\cite{wang2026normal}, and residual-guided normality learning~\cite{wang2026hunting}. 


For time-series anomaly detection, recent studies also investigate reusable pretrained detectors across domains. FAE pretrains a variational autoencoder-based generative model on large-scale time-series corpora and combines it with dilated convolutions to model complex temporal dependencies, aiming to support zero-shot anomaly detection on unseen datasets~\cite{gonzalez2025towards}. DADA further improves cross-domain transferability by introducing adaptive bottlenecks and dual adversarial decoders to handle domain divergence and diverse anomaly patterns~\cite{shentu2025towards}.

Generalist anomaly detection has also been explored in graph learning, but they are all focus on node-level anomaly detection. ARC adopts an in-context adaptation strategy that extracts dataset-specific regularities from a few normal nodes at inference time, combining feature alignment, residual encoding, and cross-attention scoring for dynamic adaptation~\cite{liu2024arc}. Similarly, AnomalyGFM develops a graph foundation model by aligning learnable normal and abnormal prototypes with residual neighborhood representations, enabling both zero-shot inference and few-shot adaptation on unseen graph datasets~\cite{qiao2025anomalygfm}. Recently, REFI-GAD introduces a fingerprint-based approach to boost the performance~\cite{liu2026rethinking}.  However, due to the fundamental differences between node-level and graph-level anomaly detection, these node-level approaches cannot be directly applied to GLAD.

Overall, existing generalist anomaly detection studies suggest that strong transferability mainly relies on robust pretrained representations together with effective test-time adaptation mechanisms, rather than dataset-specific retraining. Following this idea, our method targets a new research direction, i.e., generalist model for GLAD.

\section{Pseudo-anomalies Construction for Pre-training}
\label{Appendix:Pretraining}
Training generalist models typically depends on large volumes of labeled data~\cite{liu2022swin}. However, annotating graph-structured data demands substantial domain expertise, making it infeasible in practice~\cite{liu2023towards}. In this section, we provide a detailed version of how we construct $\mathcal{T}^\text{train}$ via  \textbf{cross-domain motif injection}. Specifically, given a batch of graphs $\mathcal{B}=\{\mathcal{G}^{\mathcal{B}}_{1}, \ldots, \mathcal{G}^{\mathcal{B}}_{B}\}$ sampled from a dataset $\mathcal{D}^{\text{train}}_\text{curr} \in \mathcal{T}^{\text{train}}$, we treat these graphs as inliers for the corresponding graph application. Pseudo-anomalies are then constructed by incorporating cross-dataset information, where each normal graph is augmented with motifs sampled from graphs drawn from other datasets to introduce anomalous information. 
Formally, for each graph $\mathcal{G}^{\mathcal{B}}_{i} \in \mathcal{B}$, we build a corresponding outlier graph $\bar{\mathcal{G}}^{\mathcal{B}}_{i}$ as a pseudo-anomaly. Let $\{\mathcal{M}_1, \ldots, \mathcal{M}_K\}$ denote the set of motifs sampled via random walks from graphs drawn from $\mathcal{T}^{\text{train}} \setminus \{\mathcal{D}^{\text{train}}_\text{curr}\}$, where each motif is defined as $\mathcal{M}_k = (\mathcal{V}^\text{motif}_k, \mathcal{E}^\text{motif}_k, \mathbf{X}^\text{motif}_k)$. We then introduce these motifs to $\mathcal{G}^{\mathcal{B}}_{i}$ by connecting a random node with the root node of the motif, resulting in the pseudo anomaly graph:
\begin{equation}
    \bar{\mathcal{G}}^{\mathcal{B}}_{i}=(\mathcal{V}^{\mathcal{B}}_{i}\cup\mathcal{V}^\text{motif}_1\cup\dots, \mathcal{E}^{\mathcal{B}}_{i}\cup\mathcal{E}^\text{motif}_1\cup\dots\cup\mathcal{E}^\text{join}, \mathbf{X}^{\mathcal{B}}_{i}\oplus \mathbf{X}^\text{motif}_1\oplus\dots), 
\end{equation}
where  $\mathcal{E}^\text{join}$  contains the edges that join the root node of motifs and the graph. Moreover, we denote the mask of injected nodes (i.e., nodes in out-domain motifs) as $\tilde{\mathbf{m}}_i \in \{0,1\}^{|\tilde{\mathcal{V}}^{\mathcal{B}}_i|}$, where 
\[
(\tilde{\mathbf{m}}_i)_j =
\begin{cases}
1, & \text{if } v_j \in \bigcup\mathcal{V}^\text{motif}_k, \\
0, & \text{otherwise}.
\end{cases}
\]
The cross-domain motif injection provides a simple yet effective way to construct negative samples. In many graph applications, the distinction between normal and anomalous instances is often determined by discriminative subgraphs that encapsulate the graph's key information~\cite {liu2023towards}. (e.g., functional groups in molecular graphs that determine chemical properties.) Injecting out-of-domain motifs perturbs the semantic consistency of inlier graphs, thereby yielding realistic pseudo-anomalies. Moreover, it also offers a fine-grained supervisory signal with injected out-of-domain motifs to guide the training of the generalist GLAD model.

\begin{algorithm}[tb]
\caption{\ourmethod: Cross-domain Motif Injection for Constructing Pseudo-anomaly}
\label{alg:cross_domain_motif_injection}

\textbf{Input:} Training task collection $\mathcal{T}$, batch of normal graph $\mathcal{B} \subset \mathcal{D}^{\text{train}}_\text{curr}\in \mathcal{T} $, number of injected motifs $K$, random walk length $l$. \\
\textbf{Output:}  A batch of pseudo-anomalies $\bar{\mathcal{B}}$ with injected node masks.

\begin{algorithmic}[1]

\STATE $\bar{\mathcal{B}} \leftarrow \emptyset$
\FOR{graph $\mathcal{G}^{\mathcal{B}}_{i}
=
(\mathcal{V}^{\mathcal{B}}_{i},
\mathcal{E}^{\mathcal{B}}_{i},
\mathbf{X}^{\mathcal{B}}_{i})
\in
\mathcal{B}_i$}
    
    \STATE Sample motifs $\{\mathcal{M}_1, ..., \mathcal{M}_K\}$ from $\mathcal{T}^{\text{train}} \setminus \{\mathcal{D}^{\text{train}}_\text{curr}\}$ with random walk of length $l$.
    \STATE $\bar{\mathcal{G}}^{\mathcal{B}}_{i}
        \leftarrow
        \mathcal{G}^{\mathcal{B}}_{i}$. 
    \FOR{$k=1,\dots,K$}
        \STATE Construct the joining edge $e^\text{join}_k $ by connecting the root node of $\mathcal{M}_k$ with a randomly selected nodes in
        $\mathcal{V}^{\mathcal{B}}_{i}$.
        \STATE Attach $\mathcal{M}_k$ to current graph:  $
    \bar{\mathcal{G}}^{\mathcal{B}}_{i}
    \leftarrow
    (
    \bar{\mathcal{V}}^{\mathcal{B}}_{i}
    \cup
    \mathcal{V}^{\text{motif}}_k,
    \bar{\mathcal{E}}^{\mathcal{B}}_{i}
    \cup
    \mathcal{E}^{\text{motif}}_k
    \cup
    \{ e^\text{join}_k \},
    \bar{\mathbf{X}}^{\mathcal{B}}_{i}
    \oplus
    \mathbf{X}^{\text{motif}}_k
    )
    $.

    \ENDFOR

    \STATE Obtain the injected node mask
    $\tilde{\mathbf{m}}_i
    \in
    \{0,1\}$.

    \STATE $
    \bar{\mathcal{B}} 
    \leftarrow 
    \bar{\mathcal{B}}  
    \cup
    \{(\bar{\mathcal{G}}^{\mathcal{B}}_{i}, \tilde{\mathbf{m}}_i)\}
    $

\ENDFOR

\STATE \textbf{return} $\bar{\mathcal{B}} $.

\end{algorithmic}
\end{algorithm}

\section{Discrepancy Design}\label{Appendix:discrepancy}
We construct a set of complementary discrepancy signals for each incoming graph, providing a holistic characterization of the incoming graph.

\noindent \textbf{Prior Discrepancy.}
We first utilize the pre-trained model to obtain the subgraph $G'$ on the test graph. Hence, the prior discrepancy is defined as:
\begin{equation}
s^{\text{prior}} = -\cos(\mathbf{z}_{\mathcal{G}'}, \mathbf{t_{\mathbf{c}}}),
\end{equation}
which estimates the difficulty of estimating the discriminative subgraph with the knowledge learned during pre-training.

\noindent \textbf{Domain Discrepancy.}
We quantify how much the graph deviates from the domain-level context. Let $\mathbf{x}_{\mathcal{G}'} = \mathrm{Readout}_{\mathcal{G}'}(\mathbf{X})$ denote the readout attribute on the discriminative subgraph. To capture complementary geometry, we consider both cosine and Euclidean distances to get discrepancy in both angular and scale differences:
\begin{equation}
s^{\text{domain}}_{\text{cos}} = -\cos(\mathbf{x}_{\mathcal{G}'}, \mathbf{c}), \quad
s^{\text{domain}}_{\text{euc}} = \|\mathbf{x}_{\mathcal{G}'} - \mathbf{c}\|_2.
\end{equation}

\noindent \textbf{Class-level Discrepancy.}
We further measure deviation from the class prototype $\mathbf{q}$ estimated from few-shot normal graphs:
\begin{equation}
\mathbf{q} = \frac{1}{|\mathcal{T}^{\text{test}}|} \sum_{\mathcal{G} \in \mathcal{T}^{\text{test}}} \mathrm{Readout}_{\mathcal{G}'}(\mathbf{H}),
\end{equation}
\begin{equation}
s^{\text{class}}_{\text{cos}} = -\cos(\mathbf{z}_{\mathcal{G}'}, \mathbf{q}), \quad
s^{\text{class}}_{\text{euc}} = \|\mathbf{z}_{\mathcal{G}'} - \mathbf{q}\|_2.
\end{equation}

\noindent \textbf{Instance-level Discrepancy.}
Beyond prototype-level statistics, we capture fine-grained relational structure via similarity distributions. Let $\mathcal{D}_{\mathcal{S}}$ denote the pairwise similarity distribution among support graphs, and $\mathcal{D}_{G}$ denote the similarity distribution between the query graph and each support instance:
\begin{equation}
\mathcal{D}_{G} = \{\cos(\mathbf{z}_{\mathcal{G}'}, \mathbf{z}_{\mathcal{G}_s})\}_{s=1}^{K}.
\end{equation}
We then define the instance-level discrepancy as the 1-D Wasserstein distance:
\begin{equation}
s^{\text{ins}} = W_1(\mathcal{D}_{\mathcal{S}}, \mathcal{D}_{G}),
\end{equation}
where the Wasserstein distance captures distributional discrepancies beyond point-wise similarity.

\section{Pseudo-anomalies Construction for Adaptation}
\label{Appendix:Finetuning}
In this section, we provide a detailed version of how we construct pseudo-anomalies from few-shot normal data supervision. Since the few-shot supervision only contains normal instances, directly learning the $\mathbf{w}$ for the scoring function is infeasible. However, as all normal instances come from the same domain, it is unsuitable apply cross-domain motif injection. While reusing the motif sampling during training can be a potential bandit, this may raise potential data leakage and privacy concerns. 

To address the aforementioned concerns, we introduce an alternative pseudo-anomaly construction way that effectively utilizes the few-shot normal graphs to generate semantically corrupted graphs and graphs with over-strong patterns, thereby supervising the contrastive learning objective with target domain-specific insights in a data-efficient manner. Specifically, to simulate anomalous graphs with out-of-domain characteristics, we build pseudo-anomaly graphs (negative outliners) by \textbf{structure perturbation}, i.e., applying structural corruption to normal graphs by shuffling edges, thereby destroying meaningful relational structures while introducing out-of-domain structure patterns.  Formally, given a normal graph $\mathcal{G} = (\mathcal{V}, \mathcal{E}, \mathbf{X})$, we generate a corrupted variant by replacing its edge set with randomly sampled node pairs. Specifically, we construct a new edge set $\tilde{\mathcal{E}}$ by sampling without replacement from the set of all possible unordered node pairs:
\begin{equation}
\tilde{\mathcal{E}} \sim \text{Uniform}\Big(\{ (u,v) \mid u,v \in \mathcal{V},\, u \neq v \}\Big), 
\quad |\tilde{\mathcal{E}}| = |\mathcal{E}|.
\end{equation}
The resulting pseudo-anomalous graph is defined as:
\begin{equation}
\tilde{\mathcal{G}}^\text{anomaly}_{\text{shuf}} = (\mathcal{V}, \tilde{\mathcal{E}}, \mathbf{X}).
\end{equation}

In addition to structural corruption, real-world anomalies may also arise from over-expressed or exaggerated in-domain patterns~\cite{kim2024rethinking}, where certain structures appear with abnormally high intensity or frequency. For example, in molecular graphs, excessive presence of certain functional groups may lead to toxicity; in social networks, while active users are typically normal, abnormally high interaction frequency can indicate spammers. Motivated by this, we conduct \textbf{sample bridging} to construct positive outliers by amplifying patterns observed in few-shot normal samples.

Specifically, given two normal graphs $\mathcal{G}_i = (\mathcal{V}_i, \mathcal{E}_i, \mathbf{X}_i)$ and $\mathcal{G}_j = (\mathcal{V}_j, \mathcal{E}_j, \mathbf{X}_j)$, we generate a bridged graph by merging them and introducing additional cross-graph connections. The resulting graph is defined as:
\begin{equation}
\tilde{\mathcal{G}}^\text{anomaly}_{\text{bridge}} = (\mathcal{V}_i \cup \mathcal{V}_j, \mathcal{E}_i \cup \mathcal{E}_j \cup \mathcal{E}_{\text{bridge}}, \mathbf{X}_i \oplus \mathbf{X}_j), 
\end{equation}
where $\mathcal{E}_{\text{bridge}}$ is the randomly sampled edge set to connect two graphs together. By bridging graphs together, this construction amplifies structural patterns beyond their normal range, thereby guilding~\ourmethod to spot any over-strong in-domain patterns.

We then utilize the constructed pseudo-anomalies with the few-shot supervisions to optimize the loss function:
\begin{equation}
\mathcal{L}_{\mathrm{adapt}}
= \mathrm{BCE}\bigl(\sigma(s), y^\text{adapt}_{\text{pseudo}}\bigr).
\end{equation}
while keeping the encoder and memory bank frozen. This preserves the transferable knowledge learned during pre-training and avoids overfitting under few-shot supervision. 
Overall, structure perturbation and sampling bridging enable \ourmethod to learn target-application specific knowledge for GLAD, thereby complementing the transferable knowledge learned during the pre-training stage with cross-domain motif injection.

\section{Algorithm}
\label{Appendix:MethodAlgorithm}
The procedure of pre-training and few-shots adaptation~\ourmethod is summarized in Algorithm~\ref{alg:pretraining} and Algorithm~\ref{alg:adaptation}, respectively.

\begin{algorithm}[tb]
\caption{\ourmethod: Generalist Pre-training}
\label{alg:pretraining}

\textbf{Input:} Training task collection $\mathcal{T}^{\text{train}}=\{\mathcal{D}^{\text{train}}_1,\dots,\mathcal{D}^{\text{train}}_N\}$, training epochs $E^\text{pre-train}$, batch size $B$, top-$K$ retrieval number, Gumbel temperature $\tau$, learning rate lr. \\
\textbf{Output:} Pre-trained \ourmethod.
\begin{algorithmic}[1]
\STATE Randomly initialize the parameters of \ourmethod.
\STATE $\mathcal{P}\leftarrow \{ (\mathcal{D}^{\text{train}}_i:\mathcal{M}^\text{pool} )\}_{i=1}^N$ (Pre-sampling motifs for each dataset to avoid computation overhead during cross-domain motif injection)
\FOR{$epoch = 1,\dots,E^\text{pre-train}$}
    \STATE $\mathcal{B}^\text{all}\leftarrow$(randomly split input datasets into batches of size B.)
    \FOR{batch $\mathcal{B} \in \mathcal{B}^\text{all}$}
        \STATE $\mathcal{L}_{\mathrm{graph}}
            \leftarrow0$
        \STATE $\mathcal{L}_{\mathrm{node}}
            \leftarrow0$
        \STATE $\mathcal{L}_{\mathrm{reg}}
            \leftarrow0$
        \STATE Retrieve top-$K$ relevant memory tokens
        $\tilde{\mathbf{T}}$
        from $\mathbf{T}$ according to cosine similarity with $\mathbf{c}$.
        \STATE Obtain pseudo-anomalies $\bar{\mathcal{B}}$ following Algorithm~\ref{alg:cross_domain_motif_injection} with the cached motifs $\mathcal{P}$
        \STATE $\tilde{\mathcal{B}}_i
        \leftarrow
        \mathcal{B}_i
        \cup
        \bar{\mathcal{B}}_i$.
        \STATE $
        \mathbf{c}
        \leftarrow
        \frac{1}{|\mathcal{B}_i|}
        \sum_{\mathcal{G}\in\mathcal{B}_i}
        \mathrm{Readout}_{\mathcal{G}}(\mathbf{X}).
        $
        \STATE $
            \mathbf{t}_{\mathbf{c}}
            \leftarrow
            \mathrm{Mean}(\tilde{\mathbf{T}}).
           $
        \FOR{graph $\tilde{\mathcal{G}}
        =
        (\tilde{\mathcal{V}},
        \tilde{\mathcal{E}},
        \tilde{\mathbf{X}})
        \in
        \tilde{\mathcal{B}}_i$}
            \STATE $\mathbf{H}
            \leftarrow
            \mathrm{GNN}(\mathbf{X},\mathbf{A}).
            $
            \STATE Obtain  $\mathcal{G}'$, $\mathbf{m}$ following Section~\ref{Method:Pretrain:subsubsec:archi}.
            \STATE $
            \mathbf{z}_{\mathcal{G}'}
            \leftarrow
            \mathrm{Readout}_{\mathcal{G}'}(\mathbf{H}).
            $
            \STATE $
            \mathcal{L}_{\mathrm{graph}}
            \leftarrow
                        \mathcal{L}_{\mathrm{graph}}
                        +
            \mathrm{BCE}
            \Bigl(
            \sigma(
            1-
            \mathrm{cos}(
            \mathbf{z}_{\mathcal{G}'},
            \mathbf{t}_{\mathbf{c}}
            )),
            \tilde{y}
            \Bigr).
            $

            \STATE         $
            \mathcal{L}_{\mathrm{node}}
            \leftarrow
            \mathcal{L}_{\mathrm{node}} 
            -
            \mathbb{E}_{u\sim\tilde{\mathbf{y}}^{(n)}_u=1}
            [\log(1-m_u)].
            $

            \STATE  $
            \mathcal{L}_{\mathrm{reg}}
            \leftarrow
            \mathcal{L}_{\mathrm{reg}}
            +
            \|\mathbf{m}\|_1.
            $

        \ENDFOR
    \STATE
    $
    \mathcal{L}_{\mathrm{pre}}
    \leftarrow
    \mathcal{L}_{\mathrm{graph}}
    +
    \lambda_n\mathcal{L}_{\mathrm{node}}
    +
    \lambda_s\mathcal{L}_{\mathrm{reg}}.
    $
    \STATE Back-propagate $ \mathcal{L}_{\mathrm{pre}}$ to update the parameters of \ourmethod with learning rate lr. 
    \ENDFOR

\ENDFOR

\STATE \textbf{return} pre-trained \ourmethod model.

\end{algorithmic}
\end{algorithm}

\begin{algorithm}[tb]
\caption{\ourmethod: Few-shot Adaptation}
\label{alg:adaptation}

\textbf{Input:} Pre-trained \ourmethod, few-shot normal set $\mathcal{S}\subset\mathcal{D}^{\text{test}}_i$, adaptation epochs $E^\text{adapt}$, learning rate lr.  \\
\textbf{Output:} Anomaly scores for graphs in $\mathcal{D}^{\text{test}}_i$.

\begin{algorithmic}[1]
\STATE Randomly initialize scoring parameter $\mathbf{w}$.
\STATE Retrieve top-$K$ relevant memory tokens $\tilde{\mathbf{T}}$ from $\mathbf{T}$ according to cosine similarity with $\mathbf{c}$.
\STATE Construct the pseudo-anomaly set $\bar{\mathcal{S}}$ $\mathcal{B}^\text{adapt}$ using  following Appendix~\ref{Appendix:Finetuning}. 

\STATE $
\mathbf{c}
\leftarrow
\frac{1}{|\mathcal{S}|}
\sum_{\mathcal{G}\in\mathcal{S}}
\mathrm{Readout}_{\mathcal{G}}(\mathbf{X}).
$

\STATE $
\mathbf{q}
\leftarrow
\frac{1}{|\mathcal{S}|}
\sum_{\mathcal{G}\in\mathcal{S}}
\mathbf{z}_{\mathcal{G}'}.
$

\FOR{$epoch=1,\dots,E^\text{adapt}$}
    \STATE $\mathcal{L}_{\mathrm{adapt}}\leftarrow0$
    \STATE Obtain $\bar{\mathcal{S}}$ following Algorithm~\ref{alg:cross_domain_motif_injection} and form fine-tuning batch
    $\tilde{\mathcal{S}}_i
    \leftarrow
    \mathcal{S}_i
    \cup
    \bar{\mathcal{S}}_i$.
    \FOR{graph $\tilde{\mathcal{G}}_i\in\tilde{\mathcal{S}}$}
        \STATE Compute the four discrepancies following Section~\ref{Method:Finetune:Multi-Dimensional Discrepancy Construction}. 
        \STATE $\mathcal{L}_{\mathrm{adapt}}\leftarrow \mathcal{L}_{\mathrm{adapt}} + \mathrm{BCE}(\sigma(\mathbf{w}^{\top}\mathbf{s}),y_i^{\mathrm{adapt}}).$
    \ENDFOR
    \STATE Update scoring parameter $\mathbf{w}$ by back-propagating $\mathcal{L}_{\mathrm{adapt}}$ with learning rate lr.
\ENDFOR

\FOR{query graph $\mathcal{G}\in\mathcal{D}^{\text{test}}_i$}
    \STATE Compute the four discrepancies following Section~\ref{Method:Finetune:Multi-Dimensional Discrepancy Construction}. 
    \STATE $s_i\leftarrow\sigma(\mathbf{w}^{\top}\mathbf{s}).$
\ENDFOR

\STATE \textbf{return} anomaly scores $\textbf{s}$.

\end{algorithmic}
\end{algorithm}

\section{Complexity Analysis}
\label{Appendix:Complexcity}

We analyze the computational complexity of each stage in \ourmethod. Let $N$ denote the total number of graphs in $\mathcal{T}^{\text{train}}$, $V$ and $E$ the average numbers of nodes and edges in a graph, respectively, $M$ the number of memory tokens, and $K$ the number of retrieved memory. We also denote the number of few-shot normal graphs as $N^\text{adapt}$. 

\noindent \textbf{During pre-training,} the time complexity of training set construction is $\mathcal{O}(N(V+E))$, where motif caching is employed to reduce computational overhead. Since the random-walk-sampled motifs are relatively small, the overall complexity is dominated by the graph merging operation, which primarily incurs copying overhead. When it comes to pre-training, for each graph, the GNN encoder requires $\mathcal{O}(E)$ for message passing. The subsequent subgraph sampling costs $\mathcal{O}(KV+E)$. In addition, it also takes $\mathcal{O}(M)$ to obtain the domain-related memory tokens for each batch. Overall, during pre-training, \ourmethod requires $\mathcal{O}(N(KV + E))$ time per training epoch, in addition to a one-time training set construction cost of $\mathcal{O}(N(V + E))$ and a per-batch overhead of $\mathcal{O}(M)$. As $K$ and $M$ are small constants, our method maintains good training efficiency and \textbf{scales comparably to training a vanilla GCN}~\cite{zhang2019graph}.

\noindent\textbf{During adaptation,}  pseudo-anomalies are constructed from $N^\text{adapt}$ few-shot normal graphs through structure perturbation and sample bridging. Structure perturbation requires $\mathcal{O}(E)$, while graph bridging costs $\mathcal{O}(V+E)$. Overall, each round of pseudo-anomaly construction requires $\mathcal{O}(N^\text{adapt}(V+E))$ as we obtain a balanced number of normal and pseudo-anomaly during training.  As inference only involves memory-guided subgraph extraction and discrepancy scoring, resulting in a complexity of $\mathcal{O}(KV+E)$. To summarize, each adaptation round incurs a time complexity of $\mathcal{O}(N^{\text{adapt}}(KV + E))$, in addition to an $\mathcal{O}(M)$ overhead for obtaining the domain-related memory tokens at the beginning of training. With the help of light-weight module design, the adaptation of \ourmethod is extremely efficient as demonstrated in Figure~\ref{Figure:Expr:Time}. 

\begin{table}[t]
\caption{Statistics of the datasets.}
\centering
\label{tab:Appendix:DatasetStatistic}
\begin{tabular}{lccc}
\toprule
Datasets & \# Graphs & Avg. Nodes & Avg. Edges \\
\midrule
\rowcolor{Gray} \multicolumn{4}{c}{\textbf{Molecular Graphs}} \\
NCI1 & 4110 & 29.87 & 64.60 \\
AIDS & 2000 & 15.59 & 47.98 \\
BZR & 405 & 35.75 & 76.72 \\
COX2 & 467 & 41.22 & 86.89 \\
DHFR & 756 & 42.43 & 89.09 \\
MUTAG & 188 & 17.93 & 57.52 \\
PTC\_MR & 344 & 14.29 & 29.38 \\
\rowcolor{Gray} \multicolumn{4}{c}{\textbf{Biochemistry Graphs}} \\
DD & 1178 & 284.32 & 1715.63 \\
ENZYMES & 600 & 32.63 & 124.27 \\
PROTEINS & 1113 & 39.06 & 145.63 \\
\rowcolor{Gray} \multicolumn{4}{c}{\textbf{Social Networks}} \\
COLLAB & 5000 & 74.49 & 4914.43 \\
IMDB-BINARY & 1000 & 19.77 & 212.84 \\
REDDIT-BINARY & 2000 & 429.63 & 1425.13 \\
\bottomrule
\end{tabular}
\end{table}

\section{Detailed Experimental Setup}
\label{Appendix:Setup}

\subsection{Description of Datasets}
In this section, we present additional details of all graph datasets used in our experiments. Table 5 reports key statistics, including the number of graphs as well as the average number of nodes and edges. All datasets are sourced from TUDataset~\cite{morris2020tudataset}, a publicly available graph benchmark repository. We follow UB-GOLD~\cite{wang2025unifying} to configure the GLAD experimental settings. Specifically, the detailed descriptions for the datasets are given as follows:

\noindent \textbf{NCI1}~\cite{wale2008comparison}: A dataset of chemical compounds screened for anti-cancer activity, where each graph corresponds to a molecular structure. It is widely used for evaluating graph learning methods in anti-cancer drug discovery tasks.

\noindent \textbf{AIDS}~\cite{morris2020tudataset}: A collection of molecular graphs representing chemical compounds tested for antiviral activity against HIV, commonly used in antiviral drug discovery research.

\noindent \textbf{BZR}~\cite{wu2018moleculenet}: A molecular dataset of benzodiazepine receptor ligands, where nodes denote atoms and edges denote chemical bonds, designed for studying molecular interactions with benzodiazepine receptors.

\noindent \textbf{COX2}~\cite{morris2020tudataset}: A dataset consisting of cyclooxygenase-2 inhibitor molecules represented as graphs, used for analyzing the structural properties of anti-inflammatory compounds.

\noindent \textbf{DHFR}~\cite{morris2020tudataset}: A collection of molecular graphs for compounds that inhibit the dihydrofolate reductase (DHFR) enzyme, important for research in cancer therapy and antibacterial drug development.

\noindent \textbf{MUTAG}~\cite{debnath1991structure}: A benchmark dataset containing 188 mutagenic aromatic and heteroaromatic nitro compounds. Each graph represents a molecule, and the classification task determines whether the compound exhibits mutagenic effects on the bacterium \textit{Salmonella typhimurium}.

\noindent \textbf{PTC-MR}~\cite{helma2001predictive}: A subset of the Predictive Toxicology Challenge (PTC) dataset, where molecular compounds are labeled according to their carcinogenicity. The suffix ``MR'' indicates experiments conducted on male rats.

\noindent \textbf{DD}~\cite{vishwanathan2010graph}: A protein-protein interaction dataset in which graphs capture the complex relationships among proteins, supporting the study of biological processes and cellular functions.

\noindent \textbf{ENZYMES}~\cite{schomburg2004brenda}: A dataset of protein tertiary structures, where each graph represents an enzyme with nodes corresponding to secondary structure elements and edges indicating spatial adjacency relationships.

\noindent \textbf{PROTEINS}~\cite{dobson2003distinguishing}: A protein graph dataset containing structural and interaction information, where nodes represent amino acids and edges capture interactions between them, facilitating the analysis of protein functions and biological interactions.

\noindent \textbf{COLLAB}~\cite{yanardag2015deep}: A scientific collaboration network dataset in which nodes correspond to researchers and edges indicate co-authorship relationships across multiple research domains.

\noindent \textbf{IMDB-BINARY}~\cite{yanardag2015deep}: A movie collaboration network dataset where nodes represent actors and edges denote co-appearances in films, enabling the analysis of collaboration patterns in the film industry.

\noindent \textbf{REDDIT-BINARY}~\cite{yanardag2015deep}: A dataset of Reddit discussion threads represented as graphs, where nodes correspond to users and edges capture interactions between them, reflecting the structural dynamics of online discussions.

As the native node and edge attributes may be unavailable or semantically incompatible across domains, we construct structure attributes, which not only addresses that issue but also unify the dimension. Such formulation can provides strong discriminative information for graph-level tasks from graph topology. Specifically, we follow the commonly used structure-based attribute construction, which consists of two components: the degree encoding $\mathbf{X}^{\mathrm{deg}}$ and the random-walk positional encoding $\mathbf{X}^{\mathrm{RW}}$. We concatenate them to obtain the structural embedding. The degree encoding is a one-hot representation of each node's degree, designed to characterize its local connectivity. We use a fixed dimensionality of 32, with degrees exceeding the supported range clipped to the final bin. To capture higher-order structural information, we additionally employ random-walk positional encoding that denotes as the probability that a k-step random walk returns to the starting point. By concatenating them together, we obtain a 64-dimensional structural embedding.

\subsection{Details of Implementation}

\noindent \textbf{Hyper-parameters. }
We tune the key hyper-parameters of \ourmethod via random search over predefined grids. Specifically, the search space includes $\lambda_n \in \{0.0, 0.1, 0.5, 1.0, 1.5, 2.0\}$, $\lambda_s \in \{0.0, 0.01, 0.05, 0.1, 0.15, 0.2\}$, and the pre-training learning rate $\text{lr}^{\text{pre-train}} \in \{1\times10^{-4}, 1\times10^{-3}\}$. For the remaining hyper-parameters, we set the number of memory tokens $M$ to 128 and the number of retrieved in-domain memory tokens $K$ to 12. For structural encoding, we set the dimensionality of both $\mathbf{X}^{\text{deg}}$ and $\mathbf{X}^{\text{RW}}$ to 32, yielding a 64-dimensional concatenated structure representation. The temperature parameter $\tau$ for Gumbel-Softmax and the hidden embedding dimension are set to 0.5 and 64, respectively. We pre-train the model for 30 epochs. For adaptation, we use a $\text{lr}^{\text{adapt}} = 1\times10^{-4}$ and train for 10 epochs. We utilize the Adam optimizer for both pre-training and fine-tuning.

\noindent \textbf{Implementation Pipeline. }
We employ a fixed set of hyperparameters to construct a generalist GLAD model across all datasets. First, all methods (including baselines and \ourmethod) are trained on the training set $\mathcal{T}^{\text{train}}$. Then, each method is adapted separately to the corresponding dataset using $\mathcal{T}^{\text{test}}$. For adaptation and few-shot fine-tuning, we ensure that the few-shot supervision is identical across all baselines and \ourmethod to guarantee a fair comparison.  During pretraining, the memory tokens are randomly initialized and jointly optimized with the model through backpropagation, while in finetuning they are frozen to avoid overfitting. 

\noindent \textbf{Evaluation Metrics. } 
Following standard practice in GAD and GLAD~\cite{ma2021comprehensive, wang2025unifying}, we evaluate performance using two widely adopted and complementary metrics: the area under the receiver operating characteristic curve (AUROC) and the area under the precision–recall curve (AUPRC). Higher AUROC and AUPRC values indicate better detection performance. We report the mean AUROC and AUPRC along with standard deviations over five independent runs.

\noindent \textbf{Hardware Configuration. } 
All the experiments are conducted on a Windows desktop equipped with an AMD Ryzen 5800X processor, 32 GB of RAM, and a single NVIDIA GeForce RTX4090 GPU with 24GB of VRAM.

\section{Additional Experimental Results}
\subsection{Main Results in AUPRC}\label{appe:exp_main}
\begin{table*}[t]
\caption{Anomaly detection performance in terms of AUPRC (in percent, mean$\pm$std). Highlighted are the results ranked \textcolor{firstcolor}{\textbf{\underline{first}}}, \textcolor{secondcolor}{\underline{second}}, and \textcolor{thirdcolor}{\underline{third}}.}
\centering
\label{tab:results_main_prc}
\resizebox{1\textwidth}{!}{
\begin{tabular}{l|cccccccccc}
\toprule
Method & AIDS & BZR & COX2 & DHFR & MUTAG & PTC\_MR & ENZYMES & PROTEINS & IMDB-B & REDDIT-B\\
\midrule

\rowcolor{Gray}
\multicolumn{11}{c}{Pre-training+Few-shot Fine-tuning} \\

  GLocalKD & $82.47{\scriptstyle\pm0.06}$ & $26.54{\scriptstyle\pm0.06}$ & \third{$25.83{\scriptstyle\pm0.09}$} & $69.52{\scriptstyle\pm0.13}$ & $87.22{\scriptstyle\pm0.05}$ & $44.04{\scriptstyle\pm0.03}$ & $81.69{\scriptstyle\pm0.02}$ & \second{$53.43{\scriptstyle\pm0.05}$} & $50.21{\scriptstyle\pm0.05}$ & \second{$69.93{\scriptstyle\pm0.13}$}\\
  CVTGAD & $76.10{\scriptstyle\pm0.00}$ & $21.15{\scriptstyle\pm0.03}$ & $23.30{\scriptstyle\pm0.01}$ & $67.41{\scriptstyle\pm0.00}$ & \third{$94.71{\scriptstyle\pm0.00}$} & $48.36{\scriptstyle\pm0.02}$ & $82.58{\scriptstyle\pm0.00}$ & $30.82{\scriptstyle\pm0.00}$ & $42.19{\scriptstyle\pm0.49}$ & OOM\\
  SIGNET & \second{$95.64{\scriptstyle\pm0.00}$} & $13.55{\scriptstyle\pm0.02}$ & $25.13{\scriptstyle\pm0.02}$ & $67.25{\scriptstyle\pm0.01}$ & $82.78{\scriptstyle\pm0.03}$ & \third{$51.26{\scriptstyle\pm0.19}$} & $81.72{\scriptstyle\pm0.00}$ & $40.01{\scriptstyle\pm0.00}$ & $52.33{\scriptstyle\pm0.01}$ & $52.30{\scriptstyle\pm0.00}$\\
  MUSE & $92.67{\scriptstyle\pm0.44}$ & $24.30{\scriptstyle\pm0.59}$ & $20.95{\scriptstyle\pm0.30}$ & $67.87{\scriptstyle\pm0.72}$ & $86.26{\scriptstyle\pm1.96}$ & $46.07{\scriptstyle\pm0.93}$ & $84.65{\scriptstyle\pm0.16}$ & $39.38{\scriptstyle\pm0.37}$ & $50.16{\scriptstyle\pm1.06}$ & $63.30{\scriptstyle\pm0.59}$\\
  GLADPro & $80.00{\scriptstyle\pm0.00}$ & $20.76{\scriptstyle\pm0.23}$ & $24.14{\scriptstyle\pm0.73}$ & $64.35{\scriptstyle\pm2.85}$ & $66.49{\scriptstyle\pm0.00}$ & $46.49{\scriptstyle\pm2.60}$ & $84.34{\scriptstyle\pm0.22}$ & $42.18{\scriptstyle\pm2.96}$ & $45.47{\scriptstyle\pm1.02}$ & $63.01{\scriptstyle\pm7.06}$\\
  HC-GLAD & OOM & OOM & OOM & OOM & OOM & OOM & OOM & OOM & OOM & OOM\\
  FANFOLD & $91.63{\scriptstyle\pm0.00}$ & $21.09{\scriptstyle\pm0.01}$ & $16.70{\scriptstyle\pm0.01}$ & $69.28{\scriptstyle\pm0.00}$ & $94.52{\scriptstyle\pm0.01}$ & $43.96{\scriptstyle\pm0.01}$ & $82.90{\scriptstyle\pm0.00}$ & $41.71{\scriptstyle\pm0.01}$ & \second{$56.12{\scriptstyle\pm0.01}$} & $34.01{\scriptstyle\pm0.00}$\\
\midrule

\rowcolor{Gray}
\multicolumn{11}{c}{Few-shot Fine-tuning Only} \\

  GLocalKD & $85.24{\scriptstyle\pm1.05}$ & \third{$26.83{\scriptstyle\pm3.24}$} & $23.23{\scriptstyle\pm0.97}$ & \second{$77.15{\scriptstyle\pm0.62}$} & \second{$94.93{\scriptstyle\pm0.24}$} & $43.21{\scriptstyle\pm0.81}$ & $81.54{\scriptstyle\pm0.86}$ & $43.71{\scriptstyle\pm3.22}$ & \third{$55.69{\scriptstyle\pm1.28}$} & $62.14{\scriptstyle\pm1.40}$\\
  CVTGAD & $89.90{\scriptstyle\pm3.71}$ & $21.82{\scriptstyle\pm6.16}$ & $20.77{\scriptstyle\pm3.09}$ & $58.70{\scriptstyle\pm3.62}$ & $83.76{\scriptstyle\pm8.41}$ & $41.03{\scriptstyle\pm2.03}$ & \second{$86.22{\scriptstyle\pm1.74}$} & $32.98{\scriptstyle\pm1.35}$ & $53.45{\scriptstyle\pm7.46}$ & OOM\\
  SIGNET & $86.33{\scriptstyle\pm2.17}$ & $18.94{\scriptstyle\pm3.97}$ & $18.45{\scriptstyle\pm1.54}$ & $52.52{\scriptstyle\pm2.12}$ & $74.93{\scriptstyle\pm11.66}$ & $45.08{\scriptstyle\pm0.62}$ & $79.56{\scriptstyle\pm0.61}$ & $48.12{\scriptstyle\pm3.36}$ & $51.78{\scriptstyle\pm3.49}$ & $48.22{\scriptstyle\pm2.48}$\\
  MUSE & \third{$95.23{\scriptstyle\pm1.85}$} & $19.21{\scriptstyle\pm0.39}$ & $18.19{\scriptstyle\pm0.59}$ & \third{$76.14{\scriptstyle\pm1.54}$} & $89.45{\scriptstyle\pm6.26}$ & $48.44{\scriptstyle\pm1.31}$ & $84.10{\scriptstyle\pm0.25}$ & $41.96{\scriptstyle\pm0.64}$ & $53.17{\scriptstyle\pm2.05}$ & \third{$68.27{\scriptstyle\pm1.44}$}\\
  GLADPro & $86.57{\scriptstyle\pm7.03}$ & \second{$28.85{\scriptstyle\pm7.18}$} & \second{$27.28{\scriptstyle\pm4.40}$} & $55.70{\scriptstyle\pm3.70}$ & $88.42{\scriptstyle\pm8.01}$ & \second{$51.52{\scriptstyle\pm1.72}$} & $85.19{\scriptstyle\pm0.45}$ & $43.32{\scriptstyle\pm5.06}$ & $44.64{\scriptstyle\pm2.32}$ & $68.24{\scriptstyle\pm3.85}$\\
  HC-GLAD & $60.55{\scriptstyle\pm0.13}$ & $18.04{\scriptstyle\pm2.13}$ & $20.33{\scriptstyle\pm0.95}$ & $65.31{\scriptstyle\pm1.67}$ & $92.86{\scriptstyle\pm1.09}$ & $43.97{\scriptstyle\pm1.27}$ & \first{$90.35{\scriptstyle\pm1.61}$} & $32.97{\scriptstyle\pm3.45}$ & $38.59{\scriptstyle\pm4.51}$ & OOM\\
  FANFOLD & $85.76{\scriptstyle\pm7.21}$ & $18.58{\scriptstyle\pm0.76}$ & $17.90{\scriptstyle\pm1.26}$ & $71.24{\scriptstyle\pm3.06}$ & $85.06{\scriptstyle\pm13.52}$ & $46.54{\scriptstyle\pm1.80}$ & $84.58{\scriptstyle\pm1.49}$ & \third{$51.46{\scriptstyle\pm4.15}$} & $52.80{\scriptstyle\pm2.68}$ & $35.58{\scriptstyle\pm1.00}$\\
\midrule
\rowcolor{Gray}
\multicolumn{11}{c}{Generalist} \\
 
  \ourmethod & \first{$98.79{\scriptstyle\pm0.01}$} & \first{$38.87{\scriptstyle\pm3.46}$} & \first{$31.64{\scriptstyle\pm1.50}$} & \first{$82.21{\scriptstyle\pm0.10}$} & \first{$95.06{\scriptstyle\pm0.13}$} & \first{$56.05{\scriptstyle\pm0.56}$} & \third{$85.47{\scriptstyle\pm0.71}$} & \first{$69.26{\scriptstyle\pm0.11}$} & \first{$74.72{\scriptstyle\pm0.64}$} & \first{$74.77{\scriptstyle\pm0.76}$}\\
\bottomrule
\end{tabular}
}
\vspace{-5mm}
\end{table*}

\begin{table*}[t]
\caption{AUROC of \ourmethod and its variants.}
\centering
\label{tab:Ablation_All_auc}
\resizebox{1\textwidth}{!}{
\begin{tabular}{l|cccccccccc}
\toprule
Variant & AIDS & BZR & COX2 & DHFR & MUTAG & PTC\_MR & ENZYMES & PROTEINS & IMDB-B & REDDIT-B\\
\midrule
\ourmethod 
& \first{$95.84{\scriptstyle\pm0.07}$} 
& \first{$63.89{\scriptstyle\pm1.18}$} 
& \first{$62.92{\scriptstyle\pm0.30}$} 
& \first{$74.96{\scriptstyle\pm0.42}$} 
& \first{$90.75{\scriptstyle\pm0.17}$} 
& \first{$60.58{\scriptstyle\pm0.47}$} 
& {$54.37{\scriptstyle\pm1.74}$} 
& {$73.63{\scriptstyle\pm0.16}$} 
& \first{$74.11{\scriptstyle\pm0.25}$} 
& \first{$73.63{\scriptstyle\pm0.20}$}\\
\midrule

\rowcolor{Gray}
\multicolumn{11}{c}{Pre-training} \\

w/o $\mathcal{L}^\text{node}$ 
& {$91.43{\scriptstyle\pm0.73}$} 
& {$57.80{\scriptstyle\pm0.73}$} 
& {$57.32{\scriptstyle\pm0.95}$} 
& {$70.98{\scriptstyle\pm0.94}$} 
& {$77.98{\scriptstyle\pm0.32}$} 
& {$58.20{\scriptstyle\pm1.69}$} 
& \first{$54.47{\scriptstyle\pm0.60}$} 
& {$66.03{\scriptstyle\pm0.19}$} 
& {$58.03{\scriptstyle\pm0.95}$} 
& {$62.77{\scriptstyle\pm0.32}$}\\

w/o $\|\mathbf{m}\|_1$ 
& {$93.82{\scriptstyle\pm0.32}$} 
& {$59.74{\scriptstyle\pm1.37}$} 
& {$58.20{\scriptstyle\pm0.17}$} 
& {$71.30{\scriptstyle\pm0.34}$} 
& {$86.32{\scriptstyle\pm0.43}$} 
& {$56.68{\scriptstyle\pm0.19}$} 
& {$53.70{\scriptstyle\pm0.29}$} 
& {$68.53{\scriptstyle\pm0.37}$} 
& {$60.06{\scriptstyle\pm2.43}$} 
& {$67.07{\scriptstyle\pm0.33}$}\\

\midrule
\rowcolor{Gray}
\multicolumn{11}{c}{Discrepancy Measures} \\

w/o Prior Disc. 
& {$93.75{\scriptstyle\pm0.16}$} 
& {$60.82{\scriptstyle\pm3.90}$} 
& {$59.03{\scriptstyle\pm1.49}$} 
& {$73.05{\scriptstyle\pm0.49}$} 
& {$88.00{\scriptstyle\pm0.59}$} 
& {$57.31{\scriptstyle\pm1.66}$} 
& {$52.54{\scriptstyle\pm1.19}$} 
& {$71.66{\scriptstyle\pm0.37}$} 
& {$71.15{\scriptstyle\pm0.99}$} 
& {$71.38{\scriptstyle\pm0.43}$}\\

w/o Domain Disc. 
& {$93.31{\scriptstyle\pm3.41}$} 
& {$57.82{\scriptstyle\pm4.22}$} 
& {$46.87{\scriptstyle\pm6.66}$} 
& {$73.77{\scriptstyle\pm0.52}$} 
& {$88.58{\scriptstyle\pm0.36}$} 
& {$46.65{\scriptstyle\pm0.67}$} 
& {$47.58{\scriptstyle\pm2.00}$} 
& {$63.76{\scriptstyle\pm1.57}$} 
& {$51.65{\scriptstyle\pm2.44}$} 
& {$65.93{\scriptstyle\pm1.78}$}\\

w/o Cls. Disc. 
& {$92.60{\scriptstyle\pm3.58}$} 
& {$52.41{\scriptstyle\pm3.05}$} 
& {$59.28{\scriptstyle\pm0.96}$} 
& {$45.63{\scriptstyle\pm1.53}$} 
& {$87.07{\scriptstyle\pm4.98}$} 
& {$55.56{\scriptstyle\pm0.55}$} 
& {$53.58{\scriptstyle\pm0.64}$} 
& {$60.20{\scriptstyle\pm1.48}$} 
& {$63.72{\scriptstyle\pm2.81}$} 
& {$61.92{\scriptstyle\pm1.45}$}\\

w/o Inst. Disc. 
& {$95.77{\scriptstyle\pm0.13}$} 
& {$56.62{\scriptstyle\pm2.39}$} 
& {$59.03{\scriptstyle\pm1.80}$} 
& {$74.25{\scriptstyle\pm0.51}$} 
& {$89.38{\scriptstyle\pm2.21}$} 
& {$58.84{\scriptstyle\pm0.74}$} 
& {$53.75{\scriptstyle\pm0.80}$} 
& \first{$73.75{\scriptstyle\pm0.14}$} 
& {$72.96{\scriptstyle\pm0.99}$} 
& {$72.16{\scriptstyle\pm1.69}$}\\

\midrule
\rowcolor{Gray}
\multicolumn{11}{c}{Scoring Module} \\

MEAN 
& {$6.15{\scriptstyle\pm0.06}$} 
& {$49.06{\scriptstyle\pm0.33}$} 
& {$56.86{\scriptstyle\pm1.27}$} 
& {$28.92{\scriptstyle\pm0.69}$} 
& {$13.34{\scriptstyle\pm0.72}$} 
& {$46.90{\scriptstyle\pm0.22}$} 
& {$46.29{\scriptstyle\pm0.21}$} 
& {$41.54{\scriptstyle\pm0.10}$} 
& {$42.00{\scriptstyle\pm0.17}$} 
& {$50.32{\scriptstyle\pm0.06}$}\\

MAX 
& {$27.63{\scriptstyle\pm0.18}$} 
& {$33.71{\scriptstyle\pm0.12}$} 
& {$57.56{\scriptstyle\pm0.01}$} 
& {$40.32{\scriptstyle\pm0.05}$} 
& {$56.57{\scriptstyle\pm0.23}$} 
& {$53.89{\scriptstyle\pm0.09}$} 
& {$44.40{\scriptstyle\pm0.08}$} 
& {$61.94{\scriptstyle\pm0.25}$} 
& {$30.12{\scriptstyle\pm0.04}$} 
& {$67.78{\scriptstyle\pm0.18}$}\\

MIN 
& {$8.44{\scriptstyle\pm0.25}$} 
& {$47.02{\scriptstyle\pm0.52}$} 
& {$58.34{\scriptstyle\pm1.84}$} 
& {$29.56{\scriptstyle\pm0.87}$} 
& {$16.36{\scriptstyle\pm1.26}$} 
& {$45.19{\scriptstyle\pm0.41}$} 
& {$46.90{\scriptstyle\pm0.53}$} 
& {$38.05{\scriptstyle\pm0.24}$} 
& {$43.57{\scriptstyle\pm0.24}$} 
& {$45.77{\scriptstyle\pm0.25}$}\\

\bottomrule
\end{tabular}
}
\end{table*}

\begin{table*}[t]
\caption{AUPRC of \ourmethod and its variants.}
\centering
\label{tab:Ablation_All_prc}
\resizebox{1\textwidth}{!}{
\begin{tabular}{l|cccccccccc}
\toprule
Variant & AIDS & BZR & COX2 & DHFR & MUTAG & PTC\_MR & ENZYMES & PROTEINS & IMDB-B & REDDIT-B\\
\midrule
\ourmethod 
& \first{$98.79{\scriptstyle\pm0.01}$} 
& \first{$38.87{\scriptstyle\pm3.46}$} 
& \first{$31.64{\scriptstyle\pm1.50}$} 
& \first{$82.21{\scriptstyle\pm0.10}$} 
& \first{$95.06{\scriptstyle\pm0.13}$} 
& \first{$56.05{\scriptstyle\pm0.56}$} 
& {$85.47{\scriptstyle\pm0.71}$} 
& \first{$69.26{\scriptstyle\pm0.11}$} 
& \first{$74.72{\scriptstyle\pm0.64}$} 
& \first{$74.77{\scriptstyle\pm0.76}$}\\

\midrule
\rowcolor{Gray}
\multicolumn{11}{c}{Pre-training} \\

w/o $\mathcal{L}^\text{node}$ 
& {$97.76{\scriptstyle\pm0.12}$} 
& {$30.72{\scriptstyle\pm0.14}$} 
& {$25.73{\scriptstyle\pm0.33}$} 
& {$80.16{\scriptstyle\pm1.33}$} 
& {$87.61{\scriptstyle\pm1.58}$} 
& {$55.29{\scriptstyle\pm1.44}$} 
& \first{$86.69{\scriptstyle\pm1.58}$} 
& {$63.80{\scriptstyle\pm0.48}$} 
& {$59.63{\scriptstyle\pm0.59}$} 
& {$67.24{\scriptstyle\pm1.74}$}\\

w/o $\|\mathbf{m}\|_1$ 
& {$98.35{\scriptstyle\pm0.14}$} 
& {$37.22{\scriptstyle\pm7.62}$} 
& {$25.87{\scriptstyle\pm0.34}$} 
& {$80.18{\scriptstyle\pm0.15}$} 
& {$92.94{\scriptstyle\pm0.13}$} 
& {$54.66{\scriptstyle\pm0.11}$} 
& {$83.76{\scriptstyle\pm0.75}$} 
& {$65.35{\scriptstyle\pm0.97}$} 
& {$55.08{\scriptstyle\pm1.31}$} 
& {$69.35{\scriptstyle\pm1.74}$}\\

\midrule
\rowcolor{Gray}
\multicolumn{11}{c}{Discrepancy Measures} \\

w/o Prior Disc. 
& {$97.79{\scriptstyle\pm0.05}$} 
& {$31.92{\scriptstyle\pm5.12}$} 
& {$27.32{\scriptstyle\pm1.23}$} 
& {$80.18{\scriptstyle\pm0.42}$} 
& {$93.77{\scriptstyle\pm0.35}$} 
& {$54.67{\scriptstyle\pm1.42}$} 
& {$82.44{\scriptstyle\pm0.60}$} 
& {$68.04{\scriptstyle\pm0.25}$} 
& {$71.85{\scriptstyle\pm1.11}$} 
& {$72.28{\scriptstyle\pm0.92}$}\\

w/o Domain Disc. 
& {$97.11{\scriptstyle\pm1.06}$} 
& {$30.19{\scriptstyle\pm8.18}$} 
& {$20.34{\scriptstyle\pm3.63}$} 
& {$81.21{\scriptstyle\pm0.35}$} 
& {$94.30{\scriptstyle\pm0.17}$} 
& {$43.49{\scriptstyle\pm0.25}$} 
& {$81.74{\scriptstyle\pm1.89}$} 
& {$59.75{\scriptstyle\pm1.27}$} 
& {$52.02{\scriptstyle\pm2.92}$} 
& {$70.22{\scriptstyle\pm1.95}$}\\

w/o Cls. Disc. 
& {$96.60{\scriptstyle\pm1.17}$} 
& {$21.32{\scriptstyle\pm2.38}$} 
& {$25.77{\scriptstyle\pm0.47}$} 
& {$63.11{\scriptstyle\pm1.57}$} 
& {$92.71{\scriptstyle\pm3.41}$} 
& {$53.54{\scriptstyle\pm0.27}$} 
& {$84.19{\scriptstyle\pm0.30}$} 
& {$54.69{\scriptstyle\pm1.92}$} 
& {$62.62{\scriptstyle\pm2.21}$} 
& {$60.04{\scriptstyle\pm1.35}$}\\

w/o Inst. Disc. 
& {$96.78{\scriptstyle\pm0.05}$} 
& {$23.59{\scriptstyle\pm1.52}$} 
& {$26.43{\scriptstyle\pm1.31}$} 
& {$81.78{\scriptstyle\pm0.11}$} 
& {$93.91{\scriptstyle\pm1.56}$} 
& {$55.22{\scriptstyle\pm0.47}$} 
& {$84.60{\scriptstyle\pm0.49}$} 
& {$69.18{\scriptstyle\pm0.08}$} 
& {$73.25{\scriptstyle\pm1.41}$} 
& {$73.69{\scriptstyle\pm1.38}$}\\

\midrule
\rowcolor{Gray}
\multicolumn{11}{c}{Scoring Module} \\

MEAN 
& {$60.97{\scriptstyle\pm0.02}$} 
& {$19.30{\scriptstyle\pm0.18}$} 
& {$24.81{\scriptstyle\pm0.94}$} 
& {$48.40{\scriptstyle\pm0.47}$} 
& {$48.75{\scriptstyle\pm0.28}$} 
& {$43.97{\scriptstyle\pm0.65}$} 
& {$80.82{\scriptstyle\pm0.19}$} 
& {$35.05{\scriptstyle\pm0.06}$} 
& {$43.70{\scriptstyle\pm0.09}$} 
& {$51.79{\scriptstyle\pm0.11}$}\\

MAX 
& {$74.09{\scriptstyle\pm0.06}$} 
& {$15.92{\scriptstyle\pm0.02}$} 
& {$24.38{\scriptstyle\pm0.01}$} 
& {$59.26{\scriptstyle\pm0.11}$} 
& {$81.51{\scriptstyle\pm0.16}$} 
& {$49.46{\scriptstyle\pm0.08}$} 
& {$79.82{\scriptstyle\pm0.06}$} 
& {$49.20{\scriptstyle\pm0.55}$} 
& {$37.89{\scriptstyle\pm0.02}$} 
& {$60.52{\scriptstyle\pm0.11}$}\\

MIN 
& {$61.64{\scriptstyle\pm0.08}$} 
& {$18.63{\scriptstyle\pm0.09}$} 
& {$26.41{\scriptstyle\pm2.57}$} 
& {$49.04{\scriptstyle\pm0.75}$} 
& {$49.45{\scriptstyle\pm0.55}$} 
& {$42.87{\scriptstyle\pm0.78}$} 
& {$81.24{\scriptstyle\pm0.22}$} 
& {$33.96{\scriptstyle\pm0.35}$} 
& {$44.73{\scriptstyle\pm0.12}$} 
& {$49.42{\scriptstyle\pm0.24}$}\\

\bottomrule
\end{tabular}
}
\end{table*}

Table~\ref{tab:results_main_prc} reports the anomaly detection performance in terms of AUPRC. Overall, \ourmethod achieves consistently strong performance across diverse datasets and demonstrates clear advantages in the generalist GLAD setting. Specifically, \ourmethod obtains the best performance on 9 out of 10 datasets, while remaining highly competitive on the remaining datasets. Notably, substantial improvements can be observed on datasets such as DHFR, IMDB-B, and REDDIT-B, indicating the effectiveness of \ourmethod in capturing transferable anomaly-related patterns across domains. 
In contrast, conventional GLAD baselines generally exhibit limited gains from pre-training and often suffer noticeable performance degradation under the few-shot fine-tuning setting. This trend is particularly evident on datasets with limited training samples, where several baselines achieve relatively low AUPRC scores or even unstable performance. These observations further validate that existing GLAD approaches heavily rely on sufficient in-domain training data and struggle to generalize across domains. 
The strong AUPRC performance of \ourmethod demonstrates that the proposed discriminative subgraph extraction and discrepancy-based adaptation strategy can effectively identify anomalous graphs while maintaining robust cross-domain generalization under limited supervision.

\subsection{More Results of Sensitivity to \#shot}\label{appe:exp_shot}

\begin{wrapfigure}{r}{0.34\textwidth}
 \centering
 \vspace{-7mm}

 \includegraphics[width=0.34\textwidth]{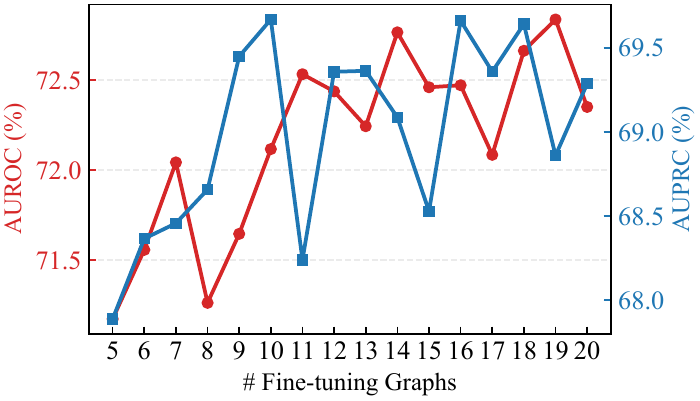}
 \caption{Sensitivity of \#shot (Science groups).}
 \label{Figure:Expr:Fewshots_sci}

 \vspace{-2mm}
\end{wrapfigure}
\begin{figure*}[t]
    \centering

    \begin{subfigure}[t]{0.19\textwidth}
        \centering
        \includegraphics[width=\textwidth]{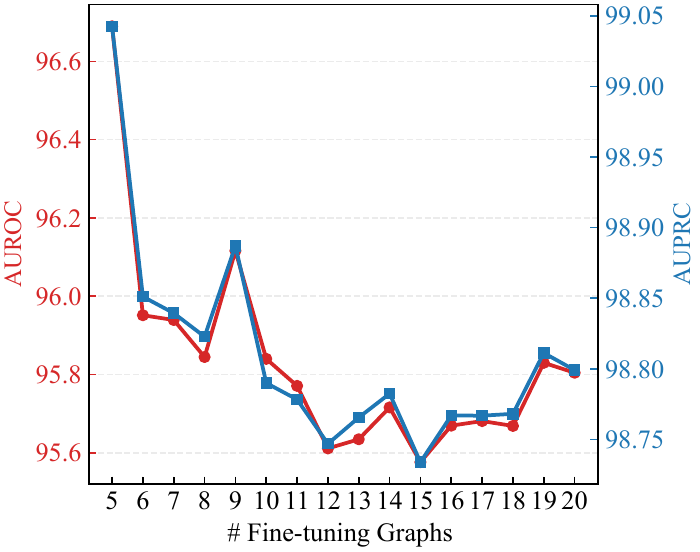}
        \caption{AIDS}
    \end{subfigure}
    \hfill
    \begin{subfigure}[t]{0.19\textwidth}
        \centering
        \includegraphics[width=\textwidth]{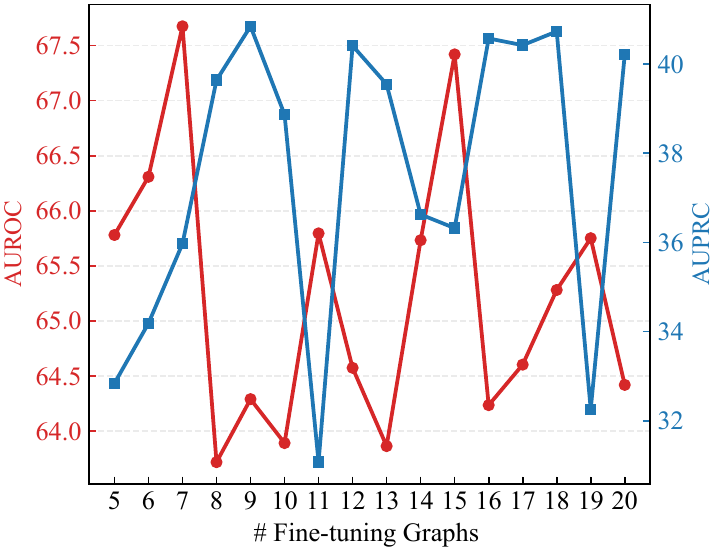}
        \caption{BZR}
    \end{subfigure}
    \hfill
    \begin{subfigure}[t]{0.19\textwidth}
        \centering
        \includegraphics[width=\textwidth]{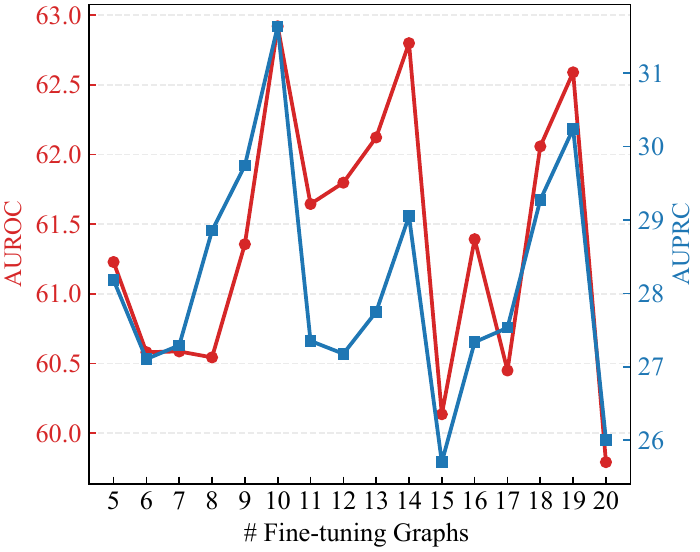}
        \caption{COX2}
    \end{subfigure}
    \hfill
    \begin{subfigure}[t]{0.19\textwidth}
        \centering
        \includegraphics[width=\textwidth]{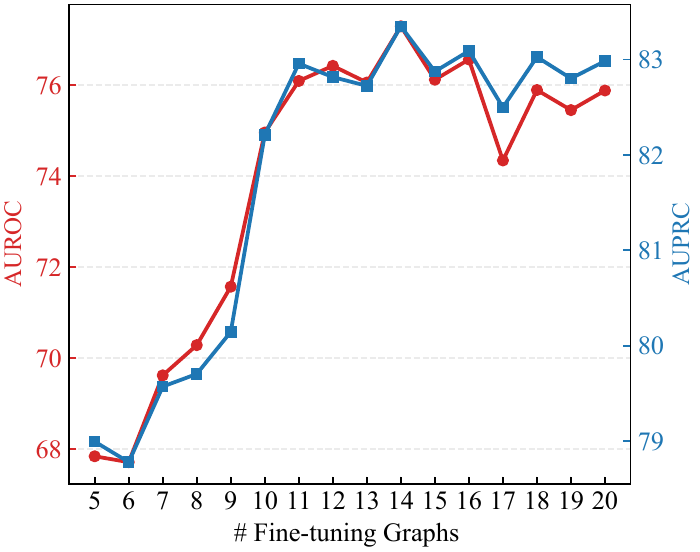}
        \caption{DHFR}
    \end{subfigure}
    \hfill
    \begin{subfigure}[t]{0.19\textwidth}
        \centering
        \includegraphics[width=\textwidth]{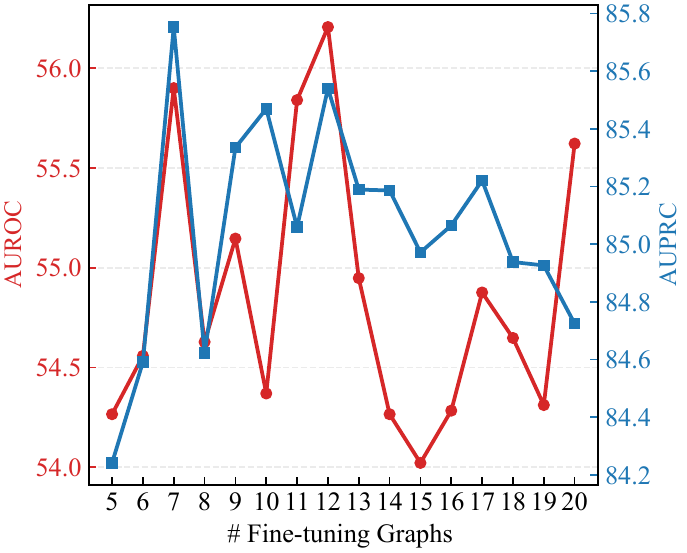}
        \caption{ENZYMES}
    \end{subfigure}

    \vspace{1mm}

    \begin{subfigure}[t]{0.19\textwidth}
        \centering
        \includegraphics[width=\textwidth]{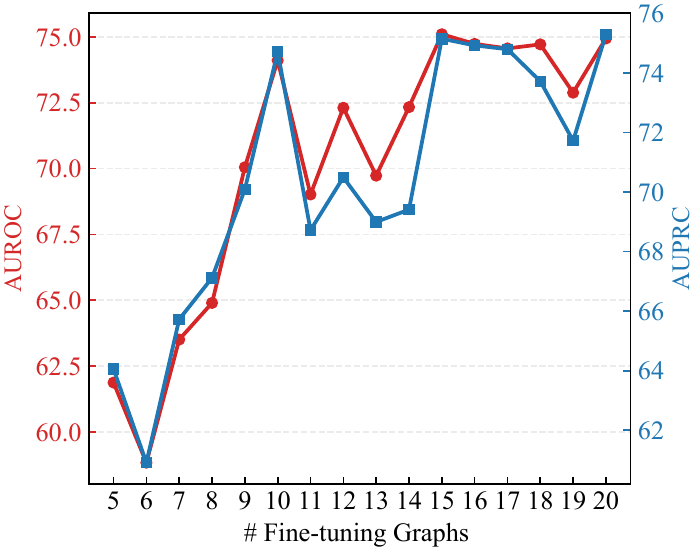}
        \caption{IMDB-B}
    \end{subfigure}
    \hfill
    \begin{subfigure}[t]{0.19\textwidth}
        \centering
        \includegraphics[width=\textwidth]{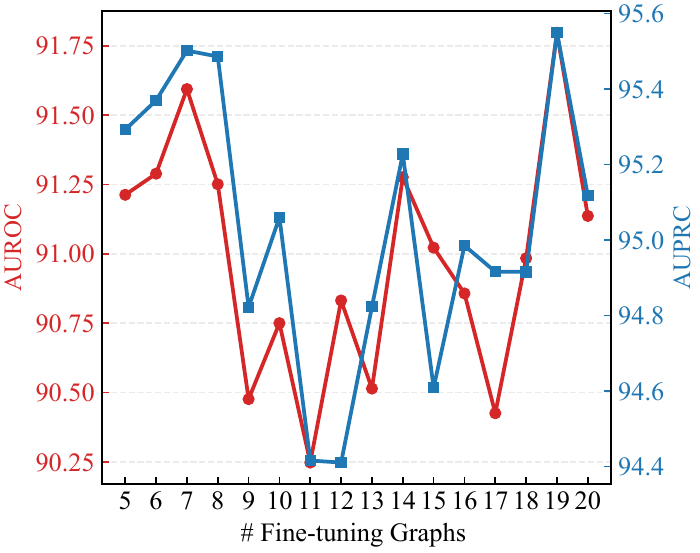}
        \caption{MUTAG}
    \end{subfigure}
    \hfill
    \begin{subfigure}[t]{0.19\textwidth}
        \centering
        \includegraphics[width=\textwidth]{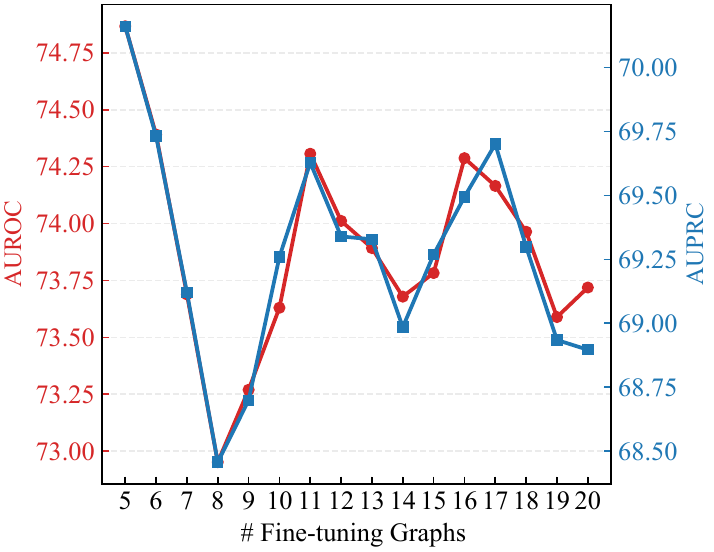}
        \caption{PROTEINS}
    \end{subfigure}
    \hfill
    \begin{subfigure}[t]{0.19\textwidth}
        \centering
        \includegraphics[width=\textwidth]{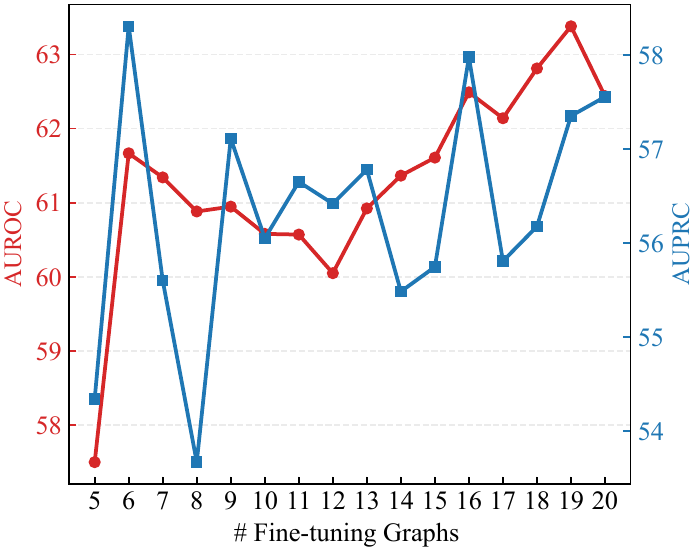}
        \caption{PTC\_MR}
    \end{subfigure}
    \hfill
    \begin{subfigure}[t]{0.19\textwidth}
        \centering
        \includegraphics[width=\textwidth]{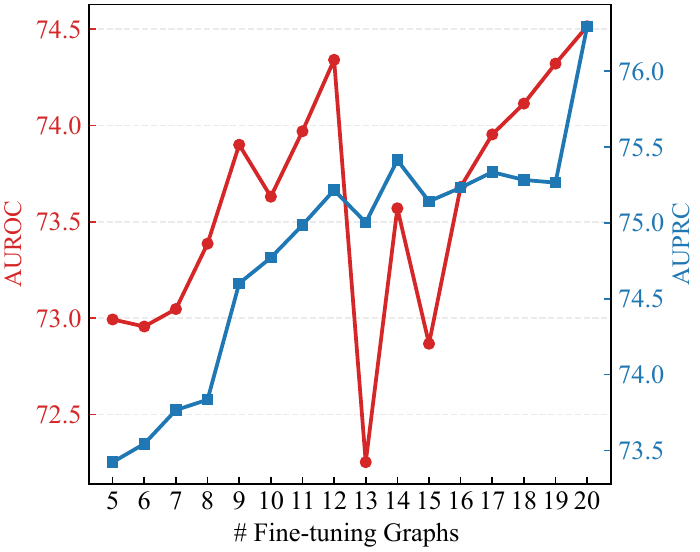}
        \caption{REDDIT-B}
    \end{subfigure}

    \caption{Performance with varying numbers of few-shot samples for adaptation.}
    \label{Figure:Expr:Fewshots_all}
\end{figure*}

Figure~\ref{Figure:Expr:Fewshots_all} presents the performance of \ourmethod with varying numbers of few-shot samples for adaptation on individual datasets, while Figure~\ref{Figure:Expr:Fewshots_sci} reports the averaged results on science graph datasets. Overall, \ourmethod demonstrates stable adaptation behavior under different few-shot settings. In most cases, increasing the number of fine-tuning graphs leads to improved AUROC and AUPRC performance, indicating that the proposed adaptation framework can effectively exploit additional normal samples when available. 

Moreover, even under extremely limited supervision, \ourmethod still achieves competitive performance on most datasets, demonstrating the strong transferability of the pre-trained backbone model. We also observe that the sensitivity to the number of few-shot samples varies across datasets. For example, datasets such as IMDB-B and DHFR exhibit relatively consistent performance improvements as more adaptation samples are introduced, while datasets such as BZR and MUTAG show larger fluctuations due to their relatively small scales and higher structural diversity. Nevertheless, the overall trends suggest that \ourmethod maintains robust adaptation capability across diverse domains and does not heavily rely on large amounts of target-domain data.

\subsection{Full Results of Ablation Study}\label{appe:exp_abl}

\noindent \textbf{Full Ablation Results. }
Tables~\ref{tab:Ablation_All_auc} and~\ref{tab:Ablation_All_prc} present the complete ablation results of \ourmethod in terms of AUROC and AUPRC, respectively. Overall, the observations are consistent with those reported in the main paper, further validating the effectiveness of each component in \ourmethod. 

For the pre-training stage, removing either the node-level pseudo anomaly supervision $\mathcal{L}_{\text{node}}$ or the sparsity regularization on $\mathbf{m}$ consistently degrades performance on most datasets. In particular, removing $\mathcal{L}_{\text{node}}$ leads to more noticeable performance drops, demonstrating the importance of fine-grained pseudo anomaly supervision for learning transferable discriminative substructures. 

For the adaptation stage, all discrepancy measures contribute positively to the final performance. Among them, the domain discrepancy and class discrepancy generally have the largest impact. Their removal causes substantial degradation on several datasets and even results in near-random performance in certain cases, highlighting the importance of capturing both domain-aware and class-level abnormality evidence. In comparison, the prior discrepancy and instance discrepancy provide complementary gains by refining the anomaly characterization from different perspectives. 

We further evaluate the effectiveness of the scoring module by replacing it with simple aggregation strategies, including mean, max, and min pooling. As shown in the tables, these heuristic strategies consistently underperform the learnable scoring module and often lead to unstable results across datasets. In some cases, they even produce performance close to or below random guessing. These observations indicate that different discrepancy signals contribute unequally across datasets, and adaptive weighting is crucial for reliable anomaly detection under few-shot adaptation settings.

\begin{figure*}
    \vspace{-2mm}
  \centering
   \includegraphics[width=\linewidth]{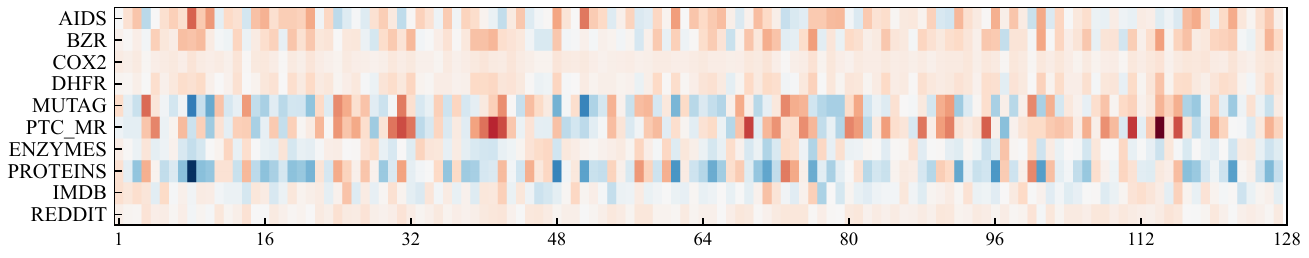}
   \vspace{-2mm}
  \caption{Visualization results of memory token importance.}\label{Figure:Expr:Tokens}
  \vspace{-2mm}
\end{figure*}

\subsection{Visualization of Memory Retrieval Weight}\label{appe:exp_vis}

As illustrated in Figure~\ref{Figure:Expr:Tokens}, token activations vary across application domains. With the help of context-guided memory retrieval, \ourmethod can selectively leverage the most relevant pre-trained knowledge for GLAD, thereby improving generalization. 

\subsection{More Analysis on Efficiency}\label{appe:efficiency}
We further provide a detailed runtime breakdown in Table~\ref{tab:camera_runtime}. As the results show, synthesizing the pseudo-graphs requires even less time than a single training epoch, demonstrating the negligible computational overhead of our graph synthesis procedure. Notably, the pretraining dataset, consisting of NCI1, DD, and COLLAB, contains 10,288 graphs, 830,144 nodes, and 26,858,668 edges in total. The reported runtime and memory consumption on this relatively large collection further demonstrate the practical efficiency and scalability of our method.

\begin{table}[t]
    \centering
    \caption{Runtime breakdown of GenGLAD in second. Pretraining is conducted on NCI1, DD, and COLLAB, while fine-tuning is evaluated on AIDS. }
    \label{tab:camera_runtime}
    \small
    \begin{tabular}{@{}p{0.78\linewidth}r@{}}
        \toprule
        Process & Time (s) \\
        \midrule
        Pretraining: pseudo-anomaly synthesis (motif pool $\mathcal{P}$) & 0.85 \\
        Pretraining: per epoch & 10.86 \\
        Pretraining: total & 325.93 \\
        \midrule
        Fine-tuning: pseudo-anomaly synthesis (pool $\mathcal{S}$) & 0.81 \\
        Fine-tuning: per epoch & 0.04 \\
        Fine-tuning: total & 0.42 \\
        \bottomrule
    \end{tabular}
\end{table}

\subsection{Quantitative Evaluation of Discriminative Subgraph}\label{appe:subgraph_quantitative}
We further provide a quantitative evaluation for the extracted discriminative subgraph on BM-MN and BM-MT. We compare GenGLAD with SIGNET, which is trained under the conventional in-domain setting. As shown in Table~\ref{tab:camera_interpretation}, despite being pretrained for cross-dataset generalization rather than optimized separately on each target dataset (like SIGNET), GenGLAD achieves interpretation performance close to SIGNET, which is trained on 500 normal graphs on BM-MN and BM-MT. Overall, these results suggest that the pretrained discriminative-subgraph extractor can identify meaningful explanatory structures across diverse anomaly patterns.

\begin{table}[t]
    \centering
    \caption{Quantitative Evaluation of Discriminative Subgraph. Results in AUROC (\%). }
    \label{tab:camera_interpretation}
    \begin{tabular}{@{}lcc@{}}
        \toprule
        Method & BM-MN & BM-MT \\
        \midrule
        SIGNET & 76.6 & 78.4 \\
        GenGLAD & 72.1 & 76.7 \\
        \bottomrule
    \end{tabular}
\end{table}


\clearpage
\section*{NeurIPS Paper Checklist}

The checklist is designed to encourage best practices for responsible machine learning research, addressing issues of reproducibility, transparency, research ethics, and societal impact. Do not remove the checklist: {\bf The papers not including the checklist will be desk rejected.} The checklist should follow the references and follow the (optional) supplemental material.  The checklist does NOT count towards the page
limit. 

Please read the checklist guidelines carefully for information on how to answer these questions. For each question in the checklist:
\begin{itemize}
    \item You should answer \answerYes{}, \answerNo{}, or \answerNA{}.
    \item \answerNA{} means either that the question is Not Applicable for that particular paper or the relevant information is Not Available.
    \item Please provide a short (1--2 sentence) justification right after your answer (even for \answerNA). 
\end{itemize}

{\bf The checklist answers are an integral part of your paper submission.} They are visible to the reviewers, area chairs, senior area chairs, and ethics reviewers. You will also be asked to include it (after eventual revisions) with the final version of your paper, and its final version will be published with the paper.

The reviewers of your paper will be asked to use the checklist as one of the factors in their evaluation. While \answerYes{} is generally preferable to \answerNo{}, it is perfectly acceptable to answer \answerNo{} provided a proper justification is given (e.g., error bars are not reported because it would be too computationally expensive'' or ``we were unable to find the license for the dataset we used''). In general, answering \answerNo{} or \answerNA{} is not grounds for rejection. While the questions are phrased in a binary way, we acknowledge that the true answer is often more nuanced, so please just use your best judgment and write a justification to elaborate. All supporting evidence can appear either in the main paper or the supplemental material, provided in appendix. If you answer \answerYes{} to a question, in the justification please point to the section(s) where related material for the question can be found.

IMPORTANT, please:
\begin{itemize}
    \item {\bf Delete this instruction block, but keep the section heading ``NeurIPS Paper Checklist"},
    \item  {\bf Keep the checklist subsection headings, questions/answers and guidelines below.}
    \item {\bf Do not modify the questions and only use the provided macros for your answers}.
\end{itemize}


\begin{enumerate}

\item {\bf Claims}
    \item[] Question: Do the main claims made in the abstract and introduction accurately reflect the paper's contributions and scope?
    \item[] Answer: \answerYes{} 
    \item[] Justification: {Our paper aims to introduce and address generalist GLAD, which is consistent with what we state in the abstract and introduction. }

\item {\bf Limitations}
    \item[] Question: Does the paper discuss the limitations of the work performed by the authors?
    \item[] Answer: \answerYes{} 
    \item[] Justification: {We mentioned limitations in the conclusion. }

\item {\bf Theory assumptions and proofs}
    \item[] Question: For each theoretical result, does the paper provide the full set of assumptions and a complete (and correct) proof?
    \item[] Answer: \answerNA{} 
    \item[] Justification: Does not include theoretical results. 

    \item {\bf Experimental result reproducibility}
    \item[] Question: Does the paper fully disclose all the information needed to reproduce the main experimental results of the paper to the extent that it affects the main claims and/or conclusions of the paper (regardless of whether the code and data are provided or not)?
    \item[] Answer: \answerYes{} 
    \item[] Justification:  We report all the hyperparameters in the appendix. The detailed algorithm can also be find in the appendix. 

\item {\bf Open access to data and code}
    \item[] Question: Does the paper provide open access to the data and code, with sufficient instructions to faithfully reproduce the main experimental results, as described in the supplemental material?
    \item[] Answer: \answerYes{} 
    \item[] Justification: We use publicly available datasets, and the source code is provided. 

\item {\bf Experimental setting/details}
    \item[] Question: Does the paper specify all the training and test details (e.g., data splits, hyperparameters, how they were chosen, type of optimizer) necessary to understand the results?
    \item[] Answer: \answerYes{} 
    \item[] Justification: We provide all hyperparameters together with range , and details of pre-training and fine-tuning dataset construction in the appendix. 

\item {\bf Experiment statistical significance}
    \item[] Question: Does the paper report error bars suitably and correctly defined or other appropriate information about the statistical significance of the experiments?
    \item[] Answer: \answerYes{} 
    \item[] Justification: Standard deviation error bar is provided. 

\item {\bf Experiments compute resources}
    \item[] Question: For each experiment, does the paper provide sufficient information on the computer resources (type of compute workers, memory, time of execution) needed to reproduce the experiments?
    \item[] Answer: \answerYes{} 
    \item[] Justification: We include the hardware configuration in the appendix. 
    
\item {\bf Code of ethics}
    \item[] Question: Does the research conducted in the paper conform, in every respect, with the NeurIPS Code of Ethics \url{https://neurips.cc/public/EthicsGuidelines}?
    \item[] Answer: \answerYes{} 
    \item[] Justification: We confirm compliance with the Code of Ethics. 

\item {\bf Broader impacts}
    \item[] Question: Does the paper discuss both potential positive societal impacts and negative societal impacts of the work performed?
    \item[] Answer: \answerYes{} 
    \item[] Justification: In the introduction, we highlight that generalist GLAD helps to safeguard diverse graph applications, including new domains with limited in-domain data. 
    
\item {\bf Safeguards}
    \item[] Question: Does the paper describe safeguards that have been put in place for responsible release of data or models that have a high risk for misuse (e.g., pre-trained language models, image generators, or scraped datasets)?
    \item[] Answer: \answerNA{} 
    \item[] Justification: The paper poses no such risk. 

\item {\bf Licenses for existing assets}
    \item[] Question: Are the creators or original owners of assets (e.g., code, data, models), used in the paper, properly credited and are the license and terms of use explicitly mentioned and properly respected?
    \item[] Answer: \answerYes{} 
    \item[] Justification: We include citations for all used datasets and baselines we compared. 

\item {\bf New assets}
    \item[] Question: Are new assets introduced in the paper well documented and is the documentation provided alongside the assets?
    \item[] Answer: \answerYes{} 
    \item[] Justification: We provide documentation for the code in READEME.md. 

\item {\bf Crowdsourcing and research with human subjects}
    \item[] Question: For crowdsourcing experiments and research with human subjects, does the paper include the full text of instructions given to participants and screenshots, if applicable, as well as details about compensation (if any)? 
    \item[] Answer: \answerNA{} 
    \item[] Justification: The paper does not involve crowdsourcing nor research with human subjects.

\item {\bf Institutional review board (IRB) approvals or equivalent for research with human subjects}
    \item[] Question: Does the paper describe potential risks incurred by study participants, whether such risks were disclosed to the subjects, and whether Institutional Review Board (IRB) approvals (or an equivalent approval/review based on the requirements of your country or institution) were obtained?
    \item[] Answer: \answerNA{} 
    \item[] Justification: The paper does not involve crowdsourcing nor research with human subjects.

\item {\bf Declaration of LLM usage}
    \item[] Question: Does the paper describe the usage of LLMs if it is an important, original, or non-standard component of the core methods in this research? Note that if the LLM is used only for writing, editing, or formatting purposes and does \emph{not} impact the core methodology, scientific rigor, or originality of the research, declaration is not required.
    \item[] Answer: \answerNA{} 
    \item[] Justification: LLM is only used for grammar and word choice. 

\end{enumerate}

\end{document}